\pdfoutput=1

\documentclass[11pt]{article}

\usepackage[final]{acl}

\usepackage{amsmath,amsfonts,bm}

\def\eqref#1{equation~\ref{#1}}

\def\1{\bm{1}}

\DeclareMathAlphabet{\mathsfit}{\encodingdefault}{\sfdefault}{m}{sl}
\SetMathAlphabet{\mathsfit}{bold}{\encodingdefault}{\sfdefault}{bx}{n}

\DeclareMathOperator*{\argmax}{arg\,max}
\DeclareMathOperator*{\argmin}{arg\,min}

\usepackage{times}
\usepackage{latexsym}

\usepackage[T1]{fontenc}

\usepackage[utf8]{inputenc}

\usepackage{microtype}

\usepackage{inconsolata}
\usepackage{microtype}
\usepackage{graphicx}
\usepackage{subfigure}
\usepackage{booktabs} 

\usepackage{hyperref}

\usepackage{amsmath}
\usepackage{amssymb}
\usepackage{mathtools}
\usepackage{amsthm}
\usepackage{mathabx}
\usepackage{soul}
\usepackage{url}
\usepackage{graphicx}
\usepackage{booktabs}
\usepackage{array}
\usepackage{comment}
\usepackage{amsfonts}
\usepackage{multirow}
\usepackage{dblfloatfix}
\usepackage{multicol}
\usepackage{xcolor}
\usepackage{upgreek}
\usepackage{siunitx}
\usepackage{natbib}
\usepackage{xspace}
\usepackage{adjustbox}
\usepackage{wrapfig}
\usepackage{bm}
\usepackage{mathrsfs}
\usepackage{amsfonts}
\usepackage{color}
\usepackage{colortbl}
\usepackage{CJKutf8}
\usepackage{booktabs}
\usepackage{array} 
\usepackage{mwe}
\usepackage{titlesec}
\usepackage{enumitem}
\usepackage{microtype}
\usepackage{fancypar}
\usepackage{extarrows}
\usepackage{array}
\usepackage{nicefrac}       
\usepackage{supertabular,enumitem}
\usepackage{hyperref}

\usepackage{epsfig}
\usepackage{changepage}
\usepackage{enumerate}
\usepackage{amsfonts,lscape,color,url}
\usepackage{tabularx}
\usepackage{adjustbox}
\usepackage{overpic}
\usepackage{tablefootnote}
\usepackage{amsmath}
\usepackage{colortbl}
\usepackage{pifont}
\sethlcolor{aligncolor}
\definecolor{aligncolor}{RGB}{207,226,245}

\usepackage{amsmath}
\usepackage{amssymb}
\usepackage{mathtools}
\usepackage{amsthm}

\usepackage[capitalize,noabbrev]{cleveref}

\theoremstyle{plain}

\theoremstyle{definition}

\theoremstyle{remark}

\newcommand{\ModelName}{Active-Prompt}

\newcommand{\highlight}[1]{\textcolor{black}{#1}}

\usepackage[textsize=tiny]{todonotes}

\DeclareSymbolFont{extraup}{U}{zavm}{m}{n}
\DeclareMathSymbol{\vardiamond}{\mathalpha}{extraup}{87}

\title{Active Prompting with Chain-of-Thought for Large Language Models}

\author{Shizhe Diao$^{\spadesuit}$, ~ Pengcheng Wang$^{\heartsuit}$, ~ \bf Yong Lin$^{\spadesuit}$,
~ \bf Rui Pan$^{\spadesuit}$,
~ \bf Xiang Liu$^{\clubsuit}$,
~ Tong Zhang$^{\vardiamond}$ \\
  $^{\spadesuit}$The Hong Kong University of Science and Technology\\
  $^{\heartsuit}$University of Toronto
    $^{\clubsuit}$The University of Hong Kong\\
    $^{\vardiamond}$University of Illinois Urbana-Champaign\\
    \texttt{\{sdiaoaa, ylindf, rpan\}@connect.ust.hk} 
    \texttt{pcheng.wang@mail.utoronto.ca} \\
    \texttt{xiang.liu@connect.hku.hk}
    \texttt{tozhang@illinois.edu} \\
  \\
}

\begin{document}
\maketitle

\def\thefootnote{\arabic{footnote}}

\begin{abstract}
The increasing scale of large language models (LLMs) brings emergent abilities to various complex tasks requiring reasoning, such as arithmetic and commonsense reasoning.
It is known that the effective design of task-specific prompts is critical for LLMs' ability to produce high-quality answers. 
In particular, an effective approach for complex question-and-answering tasks is example-based prompting with chain-of-thought (CoT) reasoning, which significantly improves the performance of LLMs. 
However, current CoT methods rely on a fixed set of human-annotated exemplars, which are not necessarily the most effective examples for different tasks.
This paper proposes a new method, \textbf{\ModelName}, to adapt LLMs to different tasks with task-specific example prompts (annotated with human-designed CoT reasoning).
For this purpose, we propose a solution to the key problem of determining which questions are the most important and helpful to annotate from a pool of task-specific queries. 
By borrowing ideas from the related problem of uncertainty-based active learning, we introduce several metrics to characterize the uncertainty so as to select the most uncertain questions for annotation.
Experimental results demonstrate the superiority of our proposed method, achieving superior performance on eight complex reasoning tasks.
Further analyses of different uncertainty metrics, pool sizes, zero-shot learning, and accuracy-uncertainty relationships demonstrate the effectiveness of our method.\footnote{Our code is available at \url{https://github.com/shizhediao/active-prompt}.}
\end{abstract}

\section{Introduction}
Large language models (LLMs)~\citep{raffel2020exploring, NEURIPS2020_1457c0d6, chowdhery2022palm, zhang2022opt, tay2022unifying, scao2022bloom, zeng2022glm, smith2022using} have achieved great success in recent years.
A typical way of applying LLMs is in-context learning~\citep{NEURIPS2020_1457c0d6} by providing a number of instructions and exemplars, which performs well on conventional language understanding and generation tasks but performs poorly on complex reasoning tasks~\citep{rae2021scaling, liang2022holistic, wei2022emergent}.
Recent prompting studies~\citep{wei2022chain, wang2022self, zhou2022least} found that elaborating the reasoning steps in the exemplars endows LLMs with good reasoning abilities, namely chain-of-thought (CoT) prompting.
However, chain-of-thought prompting depends on human engineering: it requires humans to select a few informative questions and then annotate them with CoT and answers.
The human-annotated exemplars (questions with annotated CoT and answers) are not necessarily the most effective for different tasks.
For example, the original chain-of-thought prompting~\citep{wei2022chain} crafted exemplars for eight questions, which are either randomly selected from the training set or manually composed by humans.
Due to there being a significant variance in the nature of reasoning tasks in terms of difficulty, scope, domain, and so on, we do not know what kind of question is the most worthy of annotating.
It is also not clear whether a particular set of exemplars is the best to elicit the desired information.
However, the good news is that annotating eight exemplars for different tasks is trivial. 
It costs little money and human effort.
In light of this, we identify the key problem as how to determine which questions are the most important and helpful for annotation.
We propose a solution to this problem by leveraging uncertainty and introducing a bit of human effort to annotate a small set of questions.
The annotation budget is reasonable.

\begin{figure*}[t]
    \centering
    \includegraphics[scale=0.47, trim=0 85 0 0]{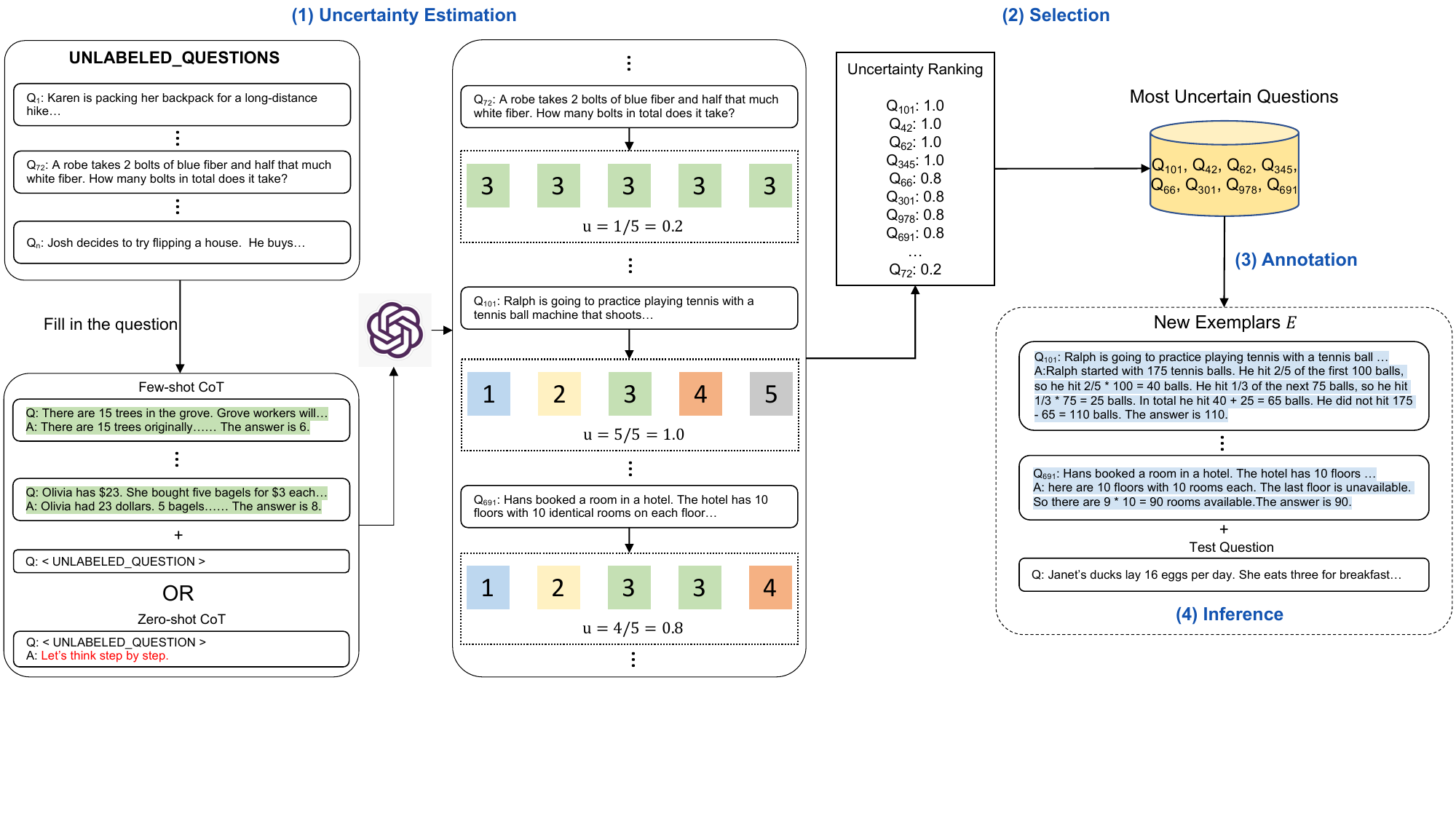}
    \caption{Illustrations of our proposed approach. There are four stages. 
    \textbf{(1) Uncertainty Estimation}: with or without a few human-written chain-of-thoughts, we query the large language model $k$ ($k=5$ in this illustration) times to generate possible answers with intermediate steps for a set of training questions. 
    Then, we calculate the uncertainty $u$ based on $k$ answers via an uncertainty metric (we use disagreement in this illustration).
    \textbf{(2) Selection}: according to the uncertainty, we select the most uncertain questions for annotation. 
    \textbf{(3) Annotation}: we involve humans to annotate the selected questions.
    \textbf{(4) Inference}: infer each question with the new annotated exemplars.}
    \label{fig:model_arch}
    \vspace{-1 em}
\end{figure*}

By borrowing ideas from the related problem of uncertainty-based active learning~\citep{gentile_active}, we introduce several metrics to characterize the uncertainty among the model's predictions on each question.
Therefore, we propose a new uncertainty-based annotation strategy that chooses a number of questions from the downstream dataset and involves humans annotating the rational chains, significantly improving the performance.
Specifically, given a dataset $D$, we first ask the model to answer it $k$ times. 
Then, we calculate the uncertainty $u$ of this model based on $k$ answers to each question.
With $u$, we select the most uncertain $n$ questions with the largest $u$ and annotate these questions by the oracle to craft new exemplars $E$.
Finally, we pre-pend $E$ to each test question following the standard recipe of chain-of-thought prompting~\citep{wei2022chain}.
The schematics of our proposed approach are illustrated in Figure~\ref{fig:model_arch}. 
There are several different ways for uncertainty estimation in the literature~\citep{settles2009active, culotta2005reducing}.
In our main experiments, we characterize the uncertainty $u$ by the disagreement and entropy of all predicted answers.
In addition, we investigate other different uncertainty metrics, like variance and self-confidence.
For self-confidence, we re-organize the generated answer with the question using a new template and then ask the model's confidence for such generation. 
In this scenario, $u$ is defined as a categorical variable from \{very confident, confident, not confident,  wrong answer\}.
It is observed that the disagreement, entropy, and variance perform similarly well, while self-confidence is not working because LLMs are prone to be over-confident. 

We conduct our experiments on eight datasets, spanning arithmetic reasoning, commonsense reasoning, and symbolic reasoning.
Experimental results demonstrate the effectiveness of our proposed method by outperforming the competitive baseline models.
Further analyses of different
uncertainty metrics, pool sizes, zero-shot learning, and accuracy-uncertainty relationship display the benefits of each proposed module and reveal their effects.
Our contributions are threefold: 
1) We propose to judiciously select the most helpful and informative questions for annotation, reducing the human engineering workload.
2) We introduce an effective uncertainty-based question selection strategy with several different uncertainty metrics.
3) Our proposed method surpasses competitive baseline models by a large margin on multiple reasoning tasks.
To the best of our knowledge, our work is the first to demonstrate the benefits of active question selection in chain-of-thought prompting for solving complex reasoning tasks.

\section{\ModelName}
The schematic illustrations of our proposed approach are illustrated in Figure~\ref{fig:model_arch}.
Given $l$ unlabeled training data $D_{tr} = \{q_1, q_2, ..., q_l\}$ and $m$ test data \highlight{$D_{te} = \{p_1, p_2, ..., p_m\}$} with each $q$ and $p$ indicating the question without any answer or reasoning steps, our goal is to annotate only $n$ questions from $D_{tr}$ as few-shot exemplars by constructing a new exemplar set $E = \{(q_1, c_1, a_1), (q_2, c_2, a_2), ..., (q_n, c_n, a_n)\}$ with reasoning steps $c$ and the answer $a$.
Then, we use $E$ to prompt the test data $D_{te}$ and obtain the predictions.
In this section, we explain how to select the $n$ most uncertain questions and annotate them.

\subsection{Uncertainty Estimation}
\label{sec:uncertainty-definition}
To select a few questions from a large dataset, we need an unsupervised method.
Previous studies~\citep{gentile_active} demonstrate that reducing the model's uncertainty helps improve the model's performance.
Therefore, we introduce the uncertainty of LLMs as a metric to select data.
In the chain-of-thought setting, we first forward the LLM $k$ times to obtain $k$ answers for each question.
Then the uncertainty of a question could be measured in different ways.
In our work, we consider four potential uncertainty metrics, described below.

\paragraph{Disagreement} 
First, we consider measuring the uncertainty using the disagreement among $k$ generated answers $A = \{a_1, a_2, ..., a_k\}$.
The disagreement is calculating the unique answers in the predictions.
The implementation is simple.
We first count the unique answers by a set operation to remove duplicate items, obtaining $h$ unique items $A = \{a_1, a_2, ..., a_h\}$.
Then, the disagreement is calculated by $u = h/k$.

\paragraph{Entropy} 
The uncertainty could also be characterized by entropy, which is calculated by
\begin{equation}
    u = \argmax_i - \sum_{j=1}^k P_\theta(a_j|q_i) \ln P_\theta(a_j|q_i),
\end{equation}
where $P_\theta(a_j|q_i)$ is the frequency of a certain predicted answer among all predictions.
A larger entropy denotes greater uncertainty in the system, and a smaller entropy denotes smaller uncertainty. 
Therefore, in complex reasoning, questions with a relatively large entropy will be selected as candidates.

\paragraph{Variance} 
We further consider variance as a kind of uncertainty metric, which we hypothesize might be more suitable for \highlight{arithmetic} answers.
\begin{equation}
    u = \argmax_i \frac{\sum_{j=1}^k(a_j - \bar{a})^2}{k - 1} \Big |_{q=q_i},
\end{equation}
where $\bar a = \frac{1}{k}\sum_{j=1}^ka_j$.
It is observed that there is a huge variation in predicted answers.
Some predicted answers are small numbers (e.g., 1), while some are large numbers (e.g., 10000).
To mitigate the domination issue of large numbers, we propose to normalize the predictions by all the mentioned numbers in the question.
For example, given a question \textit{There are $x_1$ people. Each person has $x_2$ apples. How many apples are there altogether?} and a predicted answer $\hat{y}$, we obtain $\hat{y} / (\lvert x_1 \rvert + \lvert x_2 \rvert)$ after normalization.

We first conduct a pilot study and find that disagreement-, entropy- and variance-based metrics perform competitively well, significantly outperforming self-confidence (Details are shown in Section~\ref{sec:uncertainty-metrics}).
Therefore, in our experiments, we mainly apply disagreement and entropy for our approach, which are simple to implement.

\subsection{Selection and Annotation}
After obtaining the uncertainty of each question, we can establish an uncertainty ranking according to the uncertainty of each question.
Then, we will select the top-$n$ uncertain questions for annotation.
If there are more than $n$ questions with the largest uncertainty, we will randomly select $n$ questions from them.
These $n$ questions will be annotated with rationale chains and answers by human annotators to construct new exemplars $E = \{(q_1, c_1, a_1), ..., (q_n, c_n, a_n)\}$.
$E$ will replace the initial $\hat{E}$ and we will use it for few-shot chain-to-thought prompting.

\subsection{Inference}
With the new annotated exemplars $E$, we prompt each question with them in the inference stage.
In addition, we apply self-consistency~\citep{wang2022self} to infer a question $m$ times with a temperature $T$, and then select the most consistent answer.

\section{Experimental Settings}
In this section, we describe the details of the datasets and evaluation metrics, the baseline models, and the implementation in the following three subsections. More details are included in Appendix~\ref{appendix:experimental_settings}.

\subsection{Datasets and Evaluation Metrics}
\label{sec:datasets-and-evaluation-metrics}
Following the standard evaluation settings in LLMs reasoning studies~\citep{wei2022chain}, our experiments are conducted on three types of datasets: GSM8K~\citep{cobbe2021training}, ASDiv~\citep{miao2020diverse}, SVAMP~\citep{patel2021nlp}, AQuA~\citep{ling-etal-2017-program}, SingleEq~\citep{koncel-kedziorski-etal-2016-mawps}, CSQA~\citep{talmor2019commonsenseqa}, StrategyQA~\citep{10.1162/tacl_a_00370}, and last letter concatenation~\citep{wei2022chain}.
For last letter concatenation, we test on an out-of-distribution setting, where the prompts are two letters while the test questions are four letters.
The statistics of these datasets are reported in Table~\ref{tab:dataset_statistic}.
We report the exact match accuracy as the evaluation metric.

\subsection{Baselines}
\label{sec:baselines}
In our experiments, the following four methods serve as the main baselines: Chain-of-thought (CoT)~\citep{wei2022chain}, Self-consistency (SC)~\citep{wang2022self}, Auto-CoT~\citep{zhang2022automatic}, and Random-CoT.
Random-CoT shares the same annotation process as {\ModelName}. 
The only difference is that it randomly samples questions from the training data for annotation instead of applying our proposed uncertainty metrics. 
Our experiments are mainly based on CodeX \texttt{code-davinci-002}~\citep{chen2021evaluating} for two reasons. 
First, it is the most capable model available at the time we were conducting our experiments, consistent with the observations in previous studies~\citep{wei2022chain, wang2022self, miao2020diverse}.
Second, it is free of charge in the initial limited beta period. 
In addition to \texttt{code-davinci-002}, we also test the performance with \texttt{text-davinci-002}, \texttt{text-davinci-003} and \texttt{gpt-3.5-turbo} to verify our method's effectiveness in the main experiment.
We call the APIs from OpenAI's services\footnote{\url{https://openai.com/api/}}.

\subsection{Implementation}
\label{sec:implementation}

\paragraph{Hyperparameters}
In our implementation, the model could only access the training data $D = \{X_{tr}, Y_{tr}\}$ before inference and is evaluated on the test data $D = \{X_{te}, Y_{te}\}$.
We apply the same number of exemplars as~\citet{wei2022chain}, which is 8 for GSM8K, ASDiv, SVAMP, and SingleEq, 7 for CSQA, 6 for StrategyQA, 4 for AQuA and Letter (4).
Given that some datasets (i.e., ASDiv, SVAMP, and SingleEq) only have the test split, we adopt the annotation result of GSM8K and transfer it to these datasets for inference.
The transfer details are in Table~\ref{tab:dataset_statistic}.
In the inference stage, we set temperature $T = 0.7$ and infer 40 times for each question. We then take the most consistent answer.
Unless specified, the default version of \texttt{gpt-3.5-turbo} used is \texttt{gpt-3.5-turbo-0613}.

\paragraph{Uncertainty Estimation}
At this stage, we start with a few manually annotated exemplars to help infer answers in the uncertainty estimation stage.
These annotated exemplars are directly taken from~\citet{wei2022chain}.
We call it the few-shot prompting trick to stabilize the prediction.
However, our method is not dependent on few-shot prompting, other exemplar-free methods like zero-shot prompting~\citep{kojima2022large} could be applied, and we demonstrate that it works well in Section~\ref{sec:ablation-few-shot}.
In our experiments, we limit the size of candidate instances to 1,000. 
If the size of the original training data is larger than 1,000, we only randomly sample 1,000 instances from it and consider such a subset while estimating the uncertainty.
If the size is smaller than 1,000, we will use the full data. 
We conducted the experiments with different pool sizes and found that 1,000 provides robust performance, and the performance gains of increasing the pool size would converge.
$k$ is set to 10 for all the datasets in our main experiments. 
The analysis of performance v.s. $k$ is discussed in Section~\ref{sec:ablation-few-shot}. 
The results show that with the increase in pool size, the performance continues to increase and will converge at $k$ = 10.
For the uncertainty metrics, we mainly report the performance of the disagreement-based ({\ModelName} (D)) and entropy-based ({\ModelName} (E)) methods.
Due to it having been observed that StrategyQA often ties with the maximum disagreement to be 2/2 = 1, we also take the frequency into consideration for {\ModelName} (D).

\paragraph{Annotation}
Our approach needs human annotation for a few selected questions.
The annotator is one of the co-authors and is familiar with machine learning and chain of thought prompting.
Owing to the focus of our method being the example selection rather than the annotation, the annotator did not do trial and error and conduct the minimum human engineering, referring to the previous annotation practices~\citep{wei2022chain}.
Given a question, the annotator would mainly write the reasoning steps and give the true answer to it.
The effect of different annotators and the separate effects of selection and annotation are discussed in Sections~\ref{ablation}.

\begin{table*}[t]
\centering
\footnotesize
\setlength{\tabcolsep}{0.7mm}{
\begin{tabular}{l|cccccccc|r}
\toprule
\textsc{Method} & \textsc{GSM8K} & \textsc{ASDiv} & \textsc{SVAMP} & \textsc{AQuA} & \textsc{SingleEq} & \textsc{CSQA}  & \textsc{Strategy} & \textsc{Letter (4)} & \textsc{Avg.} \\
\midrule
 Prior Best & 55.0$^a$ & 75.3$^b$ & 57.4$^c$ & 37.9$^d$ & 32.5$^e$ & 91.2$^f$ & 73.9$^g$ & - & - \\
\midrule
 \multicolumn{10}{c}{\textit{text-davinci-002}}
\\ \midrule
Auto-CoT & 47.9 & - & 69.5 & 36.5 & 87.0 & 74.4 & 65.4 & 59.7 & -  \\
CoT & 46.9 & 71.3 & 68.9 & 35.8 & 77.3 & 73.5 & 65.4 & 56.6 & 61.5 \\
SC & 58.2 & 76.9 & 78.2 & 41.8 & 87.2 & 72.9 & 70.7 & 57.6 & 67.9 \\
Random-CoT & 63.9 & 82.3 & 81.1 & 44.1 & 89.4 & 74.5 & 73.3 & 65.5 & 71.8 \\
{\ModelName} (D) & \textbf{73.2} & {83.2} & \textbf{82.7} & {48.4} & {90.6} & {76.6} & \textbf{76.9} & \textbf{67.7} & {74.9} \\
{\ModelName} (E) & {71.1} & \textbf{83.8} & {81.8} & \textbf{50.3} & \textbf{93.1} & \textbf{78.8} & \textbf{76.9} & {66.7} & \textbf{75.3} \\
\midrule
 \multicolumn{10}{c}{\textit{code-davinci-002}}
\\ \midrule
Auto-CoT & 62.8 & - & - &- &- & - & - & - & - \\
CoT & 63.1 & 80.4 & 76.4 & 45.3 & 93.1 & 77.9 & 73.2 & 70.4 & 72.5 \\
SC & 78.0 & 87.8 & 86.8 & 52.0 & 93.7 & 81.5 & 79.8 & 73.4 & 79.1 \\
Random-CoT & 78.6 & 87.1 & {88.0} & 53.1 & 94.0 & 82.1 & 79.4 & 73.3 & 79.4 \\
{\ModelName} (D) & {82.2} & {88.4} & \textbf{88.7} & {55.1} & {94.5} & \textbf{83.9} & \textbf{80.6} & {74.1} & {80.9} \\
{\ModelName} (E) & \textbf{83.4} & \textbf{89.3} & 87.5 & \textbf{57.0} & \textbf{95.5} & 82.6 & \textbf{80.6} & \textbf{76.7} & \textbf{81.6} \\
\midrule
 \multicolumn{10}{c}{\textit{\highlight{gpt-3.5-turbo-0613 (w.o. SC)}}}
\\ \midrule
CoT & 74.2 & 82.5 & 83.8 & 50.0 & 95.0 & 79.9 & 80.5 & 82.0 & 78.5 \\
{\ModelName} (D) & 77.1 & 83.6 & 85.5 & 50.0 & \textbf{96.0} & \textbf{81.5} & \textbf{82.1} & \textbf{84.0}& 80.0 \\
{\ModelName} (E) & \textbf{78.2} & \textbf{84.7} & \textbf{86.0} & \textbf{57.3} & 95.5 & 80.7 & 81.3 & \textbf{84.0}& \textbf{81.0} \\
\midrule
 \multicolumn{10}{c}{\textit{\highlight{gpt-3.5-turbo-0301} (w.o. SC)}}
\\ \midrule
CoT & 80.1 & 86.7 & 82.0 & 56.2 & 91.3 & 74.6 & 64.4 & 81.4 & 77.1 \\
{\ModelName} (D) & 83.5 & 87.4 & 83.0 & 60.6 & 93.3 & \textbf{75.9} & 70.0 & \textbf{84.0} & 79.7 \\
{\ModelName} (E) & \textbf{83.8} & \textbf{88.8} & \textbf{83.7} & \textbf{61.0} & \textbf{93.7} & 75.0 & \textbf{71.0} & \textbf{84.0} & \textbf{80.1} \\
\bottomrule
\end{tabular}
}
\caption{
The overall performance of {\ModelName}.
CoT and SC denote chain-of-thought~\citep{wei2022chain} and self-consistency~\citep{wang2022self} methods.
\textbf{Bold} denotes the best result.
$a$:~\citet{cobbe2021training}, 
$b$:~\citet{lan2022mwptoolkit}, 
$c$:~\citet{pi2022reasoning}, 
$d$:~\citet{amini-etal-2019-mathqa},
$e$:~\citet{hu2019multi},
$f$:~\citet{xu2021human},
$g$:~\citet{chowdhery2022palm}. 
Statistics for CoT and SC mostly come from the original paper, with unreported entries sourced from \textsc{DiVerSe}~\citep{li2023making}.
w.o. SC denotes that the results do not apply self-consistency, considering the cost. 
}
\label{tab:overall_performance}
\vskip -1 em
\end{table*}

\section{Experimental Results}
The experimental results are displayed in Table~\ref{tab:overall_performance}.
Overall, our model outperforms all baseline models by a large margin.
Across eight benchmark datasets, {\ModelName} (D) achieves superior results with an average of 7.0\% and 1.8\% improvement over self-consistency with \texttt{text-davinci-002} and \texttt{code-davinci-002}, respectively. 
It demonstrates the effectiveness of our proposed active selection approach. 
In this section, we discuss the results of arithmetic reasoning, commonsense and symbolic reasoning.

\paragraph{Arithmetic Reasoning:} 
{\ModelName} achieves the best performance compared with all baseline models, indicating the superiority of our method.
Compared with the competitive baseline, self-consistency, {\ModelName} (D) outperforms it by an average of 2.1\% with \texttt{code-davinci-002}.
A larger improvement is observed with \texttt{text-davinci-002}, where {\ModelName} (D) improves over self-consistency by 7.2\%.
We notice that with \texttt{code-davinci-002}, the largest improvement is observed in GSM8K (4.2\%) and AQuA (3.1\%).
One possible reason is that these two datasets do not require the transferability of CoT prompts because we can directly select and annotate the questions from their own training set.
However, ASDiv, SVAMP and SingleEq do not have training data, so we need to transfer the annotated CoT from GSM8K to them. 
This suggests that how to better transfer prompts from one task to another is considered an important future research direction.

\paragraph{Commonsense and Symbolic Reasoning:}
Consistent improvement is observed in commonsense reasoning and symbolic reasoning tasks.
{\ModelName} outperforms self-consistency across all three tasks.
Note that we test the out-of-distribution setting on Letter (4), which is more challenging, and {\ModelName} still achieves the best performance compared with all baseline models.

\begin{table}[t]
\centering
\scriptsize
\setlength{\tabcolsep}{1mm}{
\begin{tabular}{l|c|c|c|c|c}
\toprule
\textsc{Method} & GSM8K & ASDiv & SingleEq & CSQA & Letter (4) \\
\midrule
Auto-CoT & 62.8 & - & - & 74.4 & 59.7 \\
Manual-CoT & 63.1 & 80.4 & 87.5 & 73.5 & 56.6 \\
SC & 78.0 & 87.8 & 93.7 & 72.9 & 57.6 \\
Random-CoT & 78.6 & 87.1 & 94.0 & 74.5 & 65.5 \\
Zero-Shot-{\ModelName} & 82.2 & 86.7 & 94.2 & - & - \\
\midrule
{\ModelName} (D) & 82.2 & 88.4 & 94.5 & 76.6 & \textbf{67.7} \\
{\ModelName} (E) & 83.4 & 89.3 & \textbf{95.5} & \textbf{78.8} & 66.7 \\
{\ModelName} (V) & 75.2 & 86.4 & 94.0 & - & - \\
\midrule
{\ModelName}-Anno. (A) & 82.2 & 88.4 & 94.5 & 76.0 & 58.3 \\
{\ModelName}-Anno. (B) & \textbf{84.0} & \textbf{89.6} & 94.7 & 75.2 & 57.5 \\
\bottomrule
\end{tabular}
}
\caption{Ablation study on three arithmetic reasoning tasks, CSQA, and Letter (4).
Zero-Shot-{\ModelName} denotes the removal of the dependence of few-shot CoTs during uncertainty estimation.
Anno. (A) and Anno. (B) are two different annotators.
(D), (E), and (V) denote the disagreement, entropy, and variance, respectively.
\textbf{Bold} represents the best among each dataset.
The results of GSM8K, ASDiv, SingEq are obtained with  \texttt{code-davinci-002} while the results of CSQA and Letter (4) are obtained with \texttt{text-davinci-002}.
}
\label{tab:ablation_study}
\end{table}

\section{Analysis}
In this section, we further conduct several additional experiments to disclose the effects of few-shot prompts, active selection, different annotators, uncertainty metrics, pool size, and prompt engineering.
Finally, we analyze the relationship between uncertainty and accuracy, hoping to provide more explanation about how our method works.

\subsection{Ablation Study}
\label{ablation}
In this section, we reveal the impact of various modules in our proposed model design.
First, we reported the performance under the zero-shot setting by removing the dependency of a few exemplars, then explored the contributions of our proposed active example selection strategy.
In addition, we explore the effects of different annotators, different uncertainty metrics, and pool sizes.
To verify their contributions, we ablate them one by one and evaluate three downstream tasks: GSM8K, ASDiv, and SingleEq.
The results are shown in Table~\ref{tab:ablation_study}.

\paragraph{Effects of Few-Shot Prompts}
\label{sec:ablation-few-shot}
In our main experiments, we start with 4-8 manually annotated exemplars to help infer answers during the uncertainty estimation stage and demonstrate the effectiveness of our method.
These annotated exemplars are directly taken from~\citet{wei2022chain}.
However, our method is independent of the exemplars provided.
In this section, we conduct further experiments with the assumption that we do not have access to them.
Inspired by the recent research of Zero-Shot-CoT~\citep{kojima2022large}, we found it is possible to bypass the manual effort of writing the initial exemplars. 
Instead of using 4-8 human-written exemplars to generate $k$ predictions, we simply add \textit{``Let's think step by step."} and let LLMs generate the reasoning steps and the final answer. 
The results are shown in Table~\ref{tab:ablation_study} Zero-Shot-{\ModelName}, which performs competitively to {\ModelName}, demonstrating that our method is not necessarily dependent on the few-shot exemplars.

\begin{figure*}[t]
    \centering
    \includegraphics[scale=0.45, trim=90 20 0 0, clip]{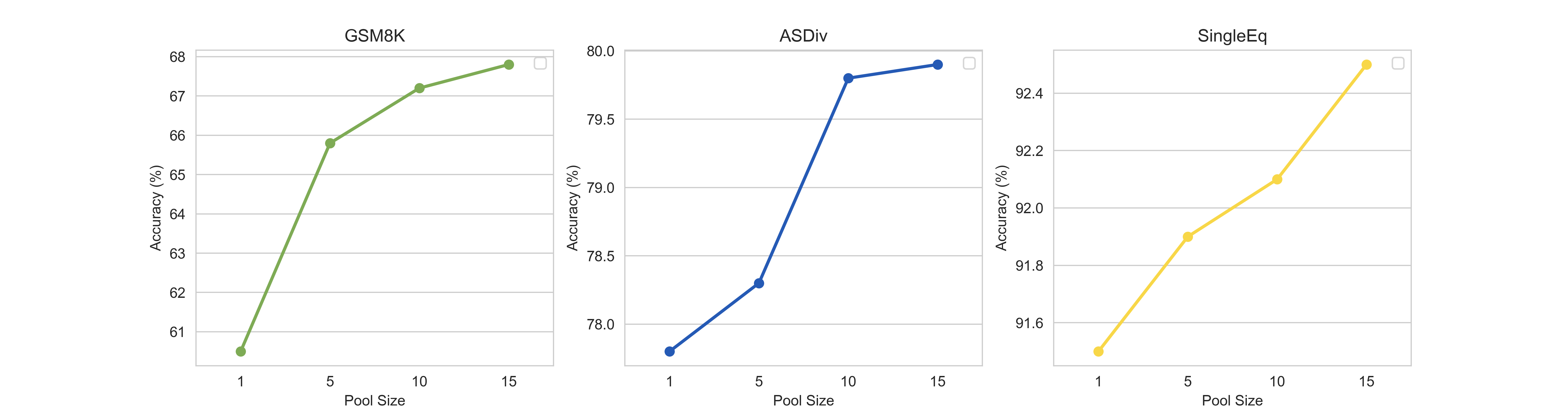}
    \caption{Comparison among the different numbers of predicted answers.}
    \label{fig:pool_size}
    \vskip -1 em
\end{figure*}

\paragraph{Effects of Active Selection}
Our main contribution is the proposal of an effective example selection strategy (namely active selection).
We replace the active selection with random selection by randomly selecting the same number of questions for annotation.
The annotation process is exactly the same as {\ModelName} with the same annotation process and annotator.
This model is called Random-CoT.
The results are shown in Table~\ref{tab:ablation_study}.
It is observed that {\ModelName} outperforms Random-CoT by a significant margin.
Random-CoT only performs comparably to another baseline model self-consistency, illustrating that our applied annotation process has no advantages, and it is the active selection strategy that leads to performance gains.
For example, on the GSM8K dataset, Random-CoT (78.6) slightly outperforms SC (78.0) while significantly underperforming {\ModelName} (82.2) by 3.6\%.
The full results of Random-CoT on all datasets are reported in Table~\ref{tab:overall_performance} with a consistent performance drop compared with {\ModelName}.

\paragraph{Effects of Annotators}
In our main experiments, we asked the annotator not to do trial and error with minimum human engineering because the focus of our method is the question selection, rather than the best possible annotation.
However, different annotators can still cause variations in the performance.
In this section, we discuss the effects of different annotators.
In addition to the annotator (annotator A), we directly use the human-annotated rationales from the GSM8K dataset (annotator B).
The results are reported in Table~\ref{tab:ablation_study}.
The results of annotators A and B are consistently better than baseline models, demonstrating the robustness of our proposed selection method.
Surprisingly, we found that directly applying the solutions provided by GSM8K outperforms our annotated rationales, suggesting that the existing annotation of GSM8K is of high quality. 
In addition, we note that human prompt engineering has two complementary components: question selection and prompt template engineering. 
The method proposed in this work provides a good solution to the first problem. It is also possible to combine this technique with human-optimized prompt templates to further improve performance.

\paragraph{Effects of Uncertainty Metrics}
\label{sec:uncertainty-metrics}
In our main experiments, we adopt disagreement and entropy as the uncertainty metric.
In addition to those, other uncertainty metrics can be incorporated.
In this section, we mainly discuss four uncertainty metrics: disagreement, entropy, variance, and self-confidence.
The definitions of the first three metrics are illustrated in Section~\ref{sec:uncertainty-definition} and the definition of self-confidence can be found in Appendix~\ref{appendix:self-confidence}.
First, we found that disagreement is not applicable to datasets with limited search space.
For example, the StrategyQA has only two labels (yes or no), and the predictions often tie in the maximum disagreement 2/2=1.
Therefore, we adopt entropy for StrategyQA.
Second, the self-confidence-based method performs badly, so we did not conduct more experiments with it.
We displayed an example of its prediction in Table~\ref{tab:example-confidence}.
We conjecture that it is because GPT-3 is prone to be over-confident, which is consistent with previous observations~\citep{si2022prompting}.
Introducing an external well-trained discriminator to evaluate confidence is a practical way, and we leave it to future work.
Last, the comparison between disagreement-, entropy- and variance-based methods are shown in Table~\ref{tab:ablation_study}.
The results illustrate that they perform competitively well on ASDiv and SingleEq, while disagreement and entropy outperform variance in GSM8K. 
Therefore, we simply choose disagreement and entropy as the primary metrics in our main experiments.


\paragraph{Effects of Pool Size}
In the first step for uncertainty estimation, we generate $k$ answers for each input question to construct a pool of predictions.
Here, $k$ affects the performance of estimating the uncertainty, further affecting the downstream task's performance.
To show the effect of the number of predicted answers, we plot the accuracy with respect to varying numbers of predicted answers (1, 5, 10, 15) in Figure~\ref{fig:pool_size} \highlight{based on \texttt{text-davinci-003}}. 
The results show that with the increase in pool size, the performance continues to increase and will converge at $k=10$.
\highlight{It is intuitive that a small $k$ may confuse the selection process, leading to ties, while a large $k$ will lead to more accurate uncertainty estimation with better performance.}

\subsection{Uncertainty Analysis}
The motivation of our proposed method is reducing the model's uncertainty to help elicit the reasoning ability of LLMs, further improving the few-shot prompting performance.
In this section, we display the relationship between uncertainty and accuracy.
In Appendix A Figure~\ref{fig:uncertainty_accuracy}, we report the uncertainty quantity and accuracy on GSM8K, ASDiv, and SingleEq.
We observe that there is a highly negative correlation between uncertainty and accuracy.
With the decrease in uncertainty, the accuracy increases, demonstrating that reducing the model's uncertainty indeed helps improve the few-shot prompting-based predictions.

\begin{table}[t]
\centering
\footnotesize
\setlength{\tabcolsep}{1.5mm}{
\begin{tabular}{l|c|c|c}
\toprule
\textsc{Method} & GSM8K & CSQA & Letter (4) \\
\midrule
TD-002 (CoT) & 46.9 & 73.5 & 56.6 \\
TD-002 $\rightarrow$ TD-002 (CoT) & 48.4 & 74.7 & 57.7 \\
\midrule
TD-002 (SC) & 58.2 & 72.9 & 57.6 \\
CD-002 $\rightarrow$ TD-002 (SC) & 73.2 & 76.6 & 67.7 \\
\midrule
TD-003 (CoT) & 61.7 & 76.2 & 70.2 \\
CD-002 $\rightarrow$ TD-003 (CoT) & 65.6 & 78.9 & 71.2 \\
\highlight{TD-003 $\rightarrow$ TD-003 (CoT)} & \highlight{67.2} & \highlight{80.8} & \highlight{73.7} \\
\bottomrule
\end{tabular}
}
\caption{
Analysis of the transferability of active exemplars.
CD-002, TD-002, TD-003 denote \texttt{code-davinci-002}, \texttt{text-davinci-002}, and \texttt{text-davinci-003}.
TD-002 (CoT), TD-002 (SC), and TD-003 (CoT)  are three baseline methods without {\ModelName}. 
TD-002 $\rightarrow$ TD-002  (CoT) denotes selecting exemplars by \texttt{text-davinci-002} and inference by \texttt{text-davinci-002}. 
CD-002 $\rightarrow$ TD-002 (SC) denotes selecting exemplars by \texttt{code-davinci-002} and inference by \texttt{text-davinci-002}. 
CD-002 $\rightarrow$ TD-003 (CoT) denotes selecting exemplars by \texttt{code-davinci-002} and inference by \texttt{text-davinci-003}.
}
\label{tab:transferability}
\vskip -1 em
\end{table}

\begin{table*}[h]
\centering
\footnotesize
\setlength{\tabcolsep}{3mm}{
\begin{tabular}{l|c|c|c|c}
\toprule
\textsc{Method} & GSM8K & AsDiv & SVAMP & SingleEq \\
\midrule
CoT (Llama2-70b-chat) & 54.8 & 73.2 & 77.4 & 84.6 \\
{\ModelName} (Llama2-70b-chat) & 57.7 & 74.5 & 82.2 & 86.8 \\
CoT (gpt-3.5-turbo) & 74.2 & 82.5 & 83.8 & 95.0 \\
{\ModelName} (gpt-3.5-turbo)  & \textbf{77.1} & \textbf{83.6} & \textbf{85.5} & \textbf{96.0} \\
\bottomrule
\end{tabular}
}
\caption{Comparison with weaker models. 
\textbf{Bold} represents the best among each dataset.
}
\label{tab:comparison_llama2}
\end{table*}

\begin{table*}[h]
\centering
\footnotesize
\setlength{\tabcolsep}{3mm}{
\begin{tabular}{l|c|c|c|c}
\toprule
\textsc{Method} & GSM8K & AsDiv & SVAMP & SingleEq \\
\midrule
gpt-3.5-turbo & 74.2 & 82.5 & 83.8 & 95.0 \\
gpt-3.5-turbo $\rightarrow$ gpt-3.5-turbo & 77.1 & 83.6 & 85.5 & 96.0 \\
Llama2-70b-chat $\rightarrow$ gpt-3.5-turbo & 78.7 & 85.9 & \highlight{84.2} & \highlight{95.4} \\
Llama2-70b-chat & 54.8 & 73.2 & 77.4 & 84.6 \\
Llama2-70b-chat $\rightarrow$ Llama2-70b-chat & 56.9 & 74.9 & 82.5 & \highlight{83.2} \\
gpt-3.5-turbo $\rightarrow$ Llama2-70b-chat & 58.9 & 74.7 & 81.2 & 86.0 \\
\bottomrule
\end{tabular}
}
\caption{Transferability between gpt-3.5-turbo and Llama models. 
}
\label{tab:transfer_chatgpt_llama}
\end{table*}

\subsection{Transferability}
In addressing the question of whether the uncertainty in selected exemplars is consistent across different models or if it originates from the specific task itself, an additional experiment was conducted. 
The experiment involves selecting exemplars using the \texttt{code-davinci-002} model and then performing inference using both \texttt{text-davinci-002} and \texttt{text-davinci-003} models. 
The underlying hypothesis is that if the uncertainty is inherent to the task, then the exemplars identified by {\ModelName} would exhibit transferability across models. 
In other words, the active exemplars identified by one model would be applicable and effective when transferred to other models. 
From the results in Table~\ref{tab:transferability}, it is observed that all three selection-based methods perform effectively.
The selected uncertain cases are related to tasks and can transfer to different models.
It indicates that the uncertainty stems from the task, and the exemplars identified by {\ModelName} demonstrate good transferability.
The results of this experiment provide insights into the nature of uncertainty in model predictions and its potential sources.

\subsection{Performance of Weaker Models}
\highlight{Our main experiments are conducted based on powerful GPT-series models.
One may wonder about the performance of weaker / smaller models, e.g., Llama-series models~\citep{touvron2023llama, touvron2023llama2}.
In this section, we investigate the effectiveness of {\ModelName} with Llama-2 models and the results are shown in Table~\ref{tab:comparison_llama2}.}
\highlight{
It is observed that our proposed {\ModelName} outperforms CoT by a large margin, demonstrating this method is still useful for weaker models.
Note that we are using the instruction-tuned version of Llama2-70b in all our experiments (i.e., Llama2-70b-chat) because it is able to understand complex chain-of-thought prompting and follow human instructions.
}

\subsection{Transferability between GPT and Llama Models}
\label{appendix:transfer_gpt_llama}
\highlight{We also investigate the transferability between GPT and Llama models. 
Because smaller Llama models perform poorly on reasoning tasks, we conduct experiments with Llama2-70b-chat. 
We conduct two types of experiments: (1) select questions by gpt-3.5-turbo and infer by Llama2-70b-chat (gpt-3.5-turbo $\rightarrow$ Llama2-70b-chat) and (2) select questions by Llama2-70b-chat and infer by gpt-3.5-turbo (Llama2-70b-chat $\rightarrow$ gpt-3.5-turbo). 
Note that we are using the 0613 version of gpt-3.5-turbo.
The results are shown in Table~\ref{tab:transfer_chatgpt_llama}.
The model before the arrow denotes the model for actively selecting questions, while the model after the arrow denotes the model for inference. 
The results demonstrate the feasibility of selecting questions with one model and then applying the selected questions to another model. 
In addition, selecting questions with larger models and applying them to smaller models results in better performance.
}

\section{Related Work}

\subsection{Chain-of-thought Prompting}
Chain-of-thought prompting elicits the reasoning abilities of large language models.
The original idea proposed by~\citet{wei2022chain} is to enrich the few-shot examples with reasoning steps, which greatly improve the performance on complex tasks.
Following~\citet{wei2022chain}, many studies improve standard CoTs in terms of self-consistency~\citep{wang2022self}, least-to-most prompting~\citep{zhou2022least}, dynamic least-to-most prompting~\citep{drozdov2022compositional}, bootstrapping~\citep{zelikman2022star}, self-training~\citep{huang2022large}, verifier~\citep{li2022advance, xu2024can}, prompt augmentation and selection~\citep{shum2023automatic}, metaheuristics~\citep{pan2023plum}, and meta-graph prompting~\citep{pan2024pomp}.
These studies greatly improve the performance based on CoT on complex tasks while they are limited to a fixed set of exemplars.
Compared with them, we propose annotating the most important task-specific questions for easy adaptation.
Auto-CoT~\citep{zhang2022automatic} clusters test questions according to the diversity and uses zero-shot prompting for answers. 
Unlike our method, it requires going through the test dataset, and our experiments show our superior performance over Auto-CoT.
Note that both diversity and uncertainty are useful for selecting the most informative questions, and they are complementary.
We consider the combination of diversity and uncertainty as a future direction.

\subsection{Active Learning}
Our work is also relevant to active learning~\citep{cohn1996active, olsson2009literature, settles2009active, rotman2022multi, lin2023optimal}, which aims to improve data labeling efficiency by finding the most helpful unlabeled data to annotate with reasonable budgets.
Recent studies~\citep{schroder2022revisiting, koksal2022meal} demonstrate the benefits of active learning-based approaches for fine-tuning large language models for classification tasks.
Following this, we incorporate max-entropy~\citep{roy2001toward}, and least confidence~\citep{culotta2005reducing} algorithms into in-context learning scenarios, and we verify the effectiveness of chain-of-thought prompting especially for complex reasoning tasks.

\section{Conclusion}
In this paper, we proposed {\ModelName} to elicit reasoning abilities in large language models (LLMs).
Inspired by the idea of annotating reasoning steps to obtain effective exemplars, we aim to select the most helpful questions for annotation judiciously instead of arbitrarily.
For this purpose, we propose an uncertainty-based active selection strategy to determine which questions are the most important and helpful to annotate from a pool of task-specific questions. 
We introduce four different strategies of uncertainty estimation for {\ModelName}: disagreement, entropy, variance, and self-confidence.
These four strategies characterize uncertainty from different perspectives, and we primarily apply disagreement and entropy.
Empirically, {\ModelName} achieved a promising performance on eight widely used datasets for arithmetic reasoning, commonsense reasoning, and symbolic reasoning.
Further analyses of different uncertainty metrics, annotators, pool sizes, zero-shot learning, and an accuracy-uncertainty relationship demonstrate the effectiveness of our method.

\section*{Limitations}
We have shown that {\ModelName} demonstrates superior performance over previous chain-of-thought prompting methods.
While exciting, there are several limitations to the current work with future opportunities.

\paragraph{Experiments with more models.} 
In our experiments, we present the complete results of \texttt{text-davinci-002} and \texttt{code-davinci-002} since \texttt{code-davinci-002} is free of charge in the initial limited beta period.
However, due to the high cost of \texttt{text-davinci-002} and \texttt{text-davinci-003}, we were only able to carry out experiments with one of them.
In addition, one promising direction is to experiment with more powerful models like GPT-4~\citep{openai2023gpt4}.
Unfortunately, conducting experiments with GPT-4 APIs is too costly.
Furthermore, we did not conduct the experiments of self-consistency with \texttt{gpt-3.5-turbo} due to the cost.
In the future, we plan to conduct experiments with GPT-4 and self-consistency experiments with \texttt{gpt-3.5-turbo} once we have more budgets.

\paragraph{Reproducibility}
In our experiments, we conduct most of the experiments with \texttt{code-davinci-002} since it is free of charge in the initial limited beta period. 
The experiments of \textit{code-davinci-002} are finished before March 2023.
However, OpenAI decided to shut off the access to \texttt{code-davinci-002}, which makes it difficult for researchers to reproduce our experiments.
However, one can access it via OpenAI's researcher access program~\footnote{\url{https://openai.com/form/researcher-access-program}} although the authors still do not have access to it.

\bibliography{custom}

\begin{thebibliography}{56}
\expandafter\ifx\csname natexlab\endcsname\relax\def\natexlab#1{#1}\fi

\bibitem[{Amini et~al.(2019)Amini, Gabriel, Lin, Koncel-Kedziorski, Choi, and Hajishirzi}]{amini-etal-2019-mathqa}
Aida Amini, Saadia Gabriel, Shanchuan Lin, Rik Koncel-Kedziorski, Yejin Choi, and Hannaneh Hajishirzi. 2019.
\newblock \href {https://doi.org/10.18653/v1/N19-1245} {{M}ath{QA}: Towards interpretable math word problem solving with operation-based formalisms}.
\newblock In \emph{Proceedings of the 2019 Conference of the North {A}merican Chapter of the Association for Computational Linguistics: Human Language Technologies, Volume 1 (Long and Short Papers)}, pages 2357--2367, Minneapolis, Minnesota. Association for Computational Linguistics.

\bibitem[{Brown et~al.(2020)Brown, Mann, Ryder, Subbiah, Kaplan, Dhariwal, Neelakantan, Shyam, Sastry, Askell, Agarwal, Herbert{-}Voss, Krueger, Henighan, Child, Ramesh, Ziegler, Wu, Winter, Hesse, Chen, Sigler, Litwin, Gray, Chess, Clark, Berner, McCandlish, Radford, Sutskever, and Amodei}]{NEURIPS2020_1457c0d6}
Tom~B. Brown, Benjamin Mann, Nick Ryder, Melanie Subbiah, Jared Kaplan, Prafulla Dhariwal, Arvind Neelakantan, Pranav Shyam, Girish Sastry, Amanda Askell, Sandhini Agarwal, Ariel Herbert{-}Voss, Gretchen Krueger, Tom Henighan, Rewon Child, Aditya Ramesh, Daniel~M. Ziegler, Jeffrey Wu, Clemens Winter, Christopher Hesse, Mark Chen, Eric Sigler, Mateusz Litwin, Scott Gray, Benjamin Chess, Jack Clark, Christopher Berner, Sam McCandlish, Alec Radford, Ilya Sutskever, and Dario Amodei. 2020.
\newblock \href {https://proceedings.neurips.cc/paper/2020/hash/1457c0d6bfcb4967418bfb8ac142f64a-Abstract.html} {Language models are few-shot learners}.
\newblock In \emph{Advances in Neural Information Processing Systems 33: Annual Conference on Neural Information Processing Systems 2020, NeurIPS 2020, December 6-12, 2020, virtual}.

\bibitem[{Chen et~al.(2021)Chen, Tworek, Jun, Yuan, Pinto, Kaplan, Edwards, Burda, Joseph, Brockman et~al.}]{chen2021evaluating}
Mark Chen, Jerry Tworek, Heewoo Jun, Qiming Yuan, Henrique Ponde de~Oliveira Pinto, Jared Kaplan, Harri Edwards, Yuri Burda, Nicholas Joseph, Greg Brockman, et~al. 2021.
\newblock Evaluating large language models trained on code.
\newblock \emph{arXiv preprint arXiv:2107.03374}.

\bibitem[{Chen et~al.(2022)Chen, Yuan, Cui, Liu, and Ji}]{chen2022close}
Yangyi Chen, Lifan Yuan, Ganqu Cui, Zhiyuan Liu, and Heng Ji. 2022.
\newblock A close look into the calibration of pre-trained language models.
\newblock \emph{arXiv preprint arXiv:2211.00151}.

\bibitem[{Chowdhery et~al.(2022)Chowdhery, Narang, Devlin, Bosma, Mishra, Roberts, Barham, Chung, Sutton, Gehrmann et~al.}]{chowdhery2022palm}
Aakanksha Chowdhery, Sharan Narang, Jacob Devlin, Maarten Bosma, Gaurav Mishra, Adam Roberts, Paul Barham, Hyung~Won Chung, Charles Sutton, Sebastian Gehrmann, et~al. 2022.
\newblock Palm: Scaling language modeling with pathways.
\newblock \emph{arXiv preprint arXiv:2204.02311}.

\bibitem[{Cobbe et~al.(2021)Cobbe, Kosaraju, Bavarian, Hilton, Nakano, Hesse, and Schulman}]{cobbe2021training}
Karl Cobbe, Vineet Kosaraju, Mohammad Bavarian, Jacob Hilton, Reiichiro Nakano, Christopher Hesse, and John Schulman. 2021.
\newblock Training verifiers to solve math word problems.
\newblock \emph{arXiv preprint arXiv:2110.14168}.

\bibitem[{Cohn et~al.(1996)Cohn, Ghahramani, and Jordan}]{cohn1996active}
David~A Cohn, Zoubin Ghahramani, and Michael~I Jordan. 1996.
\newblock Active learning with statistical models.
\newblock \emph{Journal of artificial intelligence research}, 4:129--145.

\bibitem[{Culotta and McCallum(2005)}]{culotta2005reducing}
Aron Culotta and Andrew McCallum. 2005.
\newblock Reducing labeling effort for structured prediction tasks.
\newblock In \emph{AAAI}, volume~5, pages 746--751.

\bibitem[{Drozdov et~al.(2022)Drozdov, Sch{\"a}rli, Aky{\"u}rek, Scales, Song, Chen, Bousquet, and Zhou}]{drozdov2022compositional}
Andrew Drozdov, Nathanael Sch{\"a}rli, Ekin Aky{\"u}rek, Nathan Scales, Xinying Song, Xinyun Chen, Olivier Bousquet, and Denny Zhou. 2022.
\newblock Compositional semantic parsing with large language models.
\newblock \emph{arXiv preprint arXiv:2209.15003}.

\bibitem[{Fu et~al.(2022)Fu, Peng, Sabharwal, Clark, and Khot}]{fu2022complexity}
Yao Fu, Hao Peng, Ashish Sabharwal, Peter Clark, and Tushar Khot. 2022.
\newblock Complexity-based prompting for multi-step reasoning.
\newblock \emph{arXiv preprint arXiv:2210.00720}.

\bibitem[{Gentile et~al.(2022)Gentile, Wang, and Zhang}]{gentile_active}
Claudio Gentile, Zhilei Wang, and Tong Zhang. 2022.
\newblock \href {https://doi.org/10.48550/ARXIV.2202.05448} {Fast rates in pool-based batch active learning}.

\bibitem[{Geva et~al.(2021)Geva, Khashabi, Segal, Khot, Roth, and Berant}]{10.1162/tacl_a_00370}
Mor Geva, Daniel Khashabi, Elad Segal, Tushar Khot, Dan Roth, and Jonathan Berant. 2021.
\newblock \href {https://doi.org/10.1162/tacl_a_00370} {{Did Aristotle Use a Laptop? A Question Answering Benchmark with Implicit Reasoning Strategies}}.
\newblock \emph{Transactions of the Association for Computational Linguistics}, 9:346--361.

\bibitem[{Guo et~al.(2017)Guo, Pleiss, Sun, and Weinberger}]{guo2017calibration}
Chuan Guo, Geoff Pleiss, Yu~Sun, and Kilian~Q Weinberger. 2017.
\newblock On calibration of modern neural networks.
\newblock In \emph{International conference on machine learning}, pages 1321--1330. PMLR.

\bibitem[{Hu et~al.(2019)Hu, Peng, Huang, and Li}]{hu2019multi}
Minghao Hu, Yuxing Peng, Zhen Huang, and Dongsheng Li. 2019.
\newblock A multi-type multi-span network for reading comprehension that requires discrete reasoning.
\newblock In \emph{Proceedings of the 2019 Conference on Empirical Methods in Natural Language Processing and the 9th International Joint Conference on Natural Language Processing (EMNLP-IJCNLP)}, pages 1596--1606.

\bibitem[{Huang et~al.(2022)Huang, Gu, Hou, Wu, Wang, Yu, and Han}]{huang2022large}
Jiaxin Huang, Shixiang~Shane Gu, Le~Hou, Yuexin Wu, Xuezhi Wang, Hongkun Yu, and Jiawei Han. 2022.
\newblock Large language models can self-improve.
\newblock \emph{arXiv preprint arXiv:2210.11610}.

\bibitem[{Kojima et~al.(2022)Kojima, Gu, Reid, Matsuo, and Iwasawa}]{kojima2022large}
Takeshi Kojima, Shixiang~Shane Gu, Machel Reid, Yutaka Matsuo, and Yusuke Iwasawa. 2022.
\newblock Large language models are zero-shot reasoners.
\newblock \emph{arXiv preprint arXiv:2205.11916}.

\bibitem[{K{\"o}ksal et~al.(2022)K{\"o}ksal, Schick, and Sch{\"u}tze}]{koksal2022meal}
Abdullatif K{\"o}ksal, Timo Schick, and Hinrich Sch{\"u}tze. 2022.
\newblock Meal: Stable and active learning for few-shot prompting.
\newblock \emph{arXiv preprint arXiv:2211.08358}.

\bibitem[{Koncel-Kedziorski et~al.(2016)Koncel-Kedziorski, Roy, Amini, Kushman, and Hajishirzi}]{koncel-kedziorski-etal-2016-mawps}
Rik Koncel-Kedziorski, Subhro Roy, Aida Amini, Nate Kushman, and Hannaneh Hajishirzi. 2016.
\newblock \href {https://doi.org/10.18653/v1/N16-1136} {{MAWPS}: A math word problem repository}.
\newblock In \emph{Proceedings of the 2016 Conference of the North {A}merican Chapter of the Association for Computational Linguistics: Human Language Technologies}, pages 1152--1157, San Diego, California. Association for Computational Linguistics.

\bibitem[{Kong et~al.(2020)Kong, Jiang, Zhuang, Lyu, Zhao, and Zhang}]{kong2020calibrated}
Lingkai Kong, Haoming Jiang, Yuchen Zhuang, Jie Lyu, Tuo Zhao, and Chao Zhang. 2020.
\newblock Calibrated language model fine-tuning for in-and out-of-distribution data.
\newblock \emph{arXiv preprint arXiv:2010.11506}.

\bibitem[{Lan et~al.(2022)Lan, Wang, Zhang, Lan, Dai, Wang, Zhang, and Lim}]{lan2022mwptoolkit}
Yihuai Lan, Lei Wang, Qiyuan Zhang, Yunshi Lan, Bing~Tian Dai, Yan Wang, Dongxiang Zhang, and Ee-Peng Lim. 2022.
\newblock Mwptoolkit: an open-source framework for deep learning-based math word problem solvers.
\newblock In \emph{Proceedings of the AAAI Conference on Artificial Intelligence}, volume~36, pages 13188--13190.

\bibitem[{Li et~al.(2022)Li, Lin, Zhang, Fu, Chen, Lou, and Chen}]{li2022advance}
Yifei Li, Zeqi Lin, Shizhuo Zhang, Qiang Fu, Bei Chen, Jian-Guang Lou, and Weizhu Chen. 2022.
\newblock On the advance of making language models better reasoners.
\newblock \emph{arXiv preprint arXiv:2206.02336}.

\bibitem[{Li et~al.(2023)Li, Lin, Zhang, Fu, Chen, Lou, and Chen}]{li2023making}
Yifei Li, Zeqi Lin, Shizhuo Zhang, Qiang Fu, Bei Chen, Jian-Guang Lou, and Weizhu Chen. 2023.
\newblock Making language models better reasoners with step-aware verifier.
\newblock In \emph{Proceedings of the 61st Annual Meeting of the Association for Computational Linguistics (Volume 1: Long Papers)}, pages 5315--5333.

\bibitem[{Liang et~al.(2022)Liang, Bommasani, Lee, Tsipras, Soylu, Yasunaga, Zhang, Narayanan, Wu, Kumar et~al.}]{liang2022holistic}
Percy Liang, Rishi Bommasani, Tony Lee, Dimitris Tsipras, Dilara Soylu, Michihiro Yasunaga, Yian Zhang, Deepak Narayanan, Yuhuai Wu, Ananya Kumar, et~al. 2022.
\newblock Holistic evaluation of language models.
\newblock \emph{arXiv preprint arXiv:2211.09110}.

\bibitem[{Lin et~al.(2023)Lin, Liu, Ye, Lian, Yao, and Zhang}]{lin2023optimal}
Yong Lin, Chen Liu, Chenlu Ye, Qing Lian, Yuan Yao, and Tong Zhang. 2023.
\newblock Optimal sample selection through uncertainty estimation and its application in deep learning.
\newblock \emph{arXiv preprint arXiv:2309.02476}.

\bibitem[{Ling et~al.(2017)Ling, Yogatama, Dyer, and Blunsom}]{ling-etal-2017-program}
Wang Ling, Dani Yogatama, Chris Dyer, and Phil Blunsom. 2017.
\newblock \href {https://doi.org/10.18653/v1/P17-1015} {Program induction by rationale generation: Learning to solve and explain algebraic word problems}.
\newblock In \emph{Proceedings of the 55th Annual Meeting of the Association for Computational Linguistics (Volume 1: Long Papers)}, pages 158--167, Vancouver, Canada. Association for Computational Linguistics.

\bibitem[{Miao et~al.(2020)Miao, Liang, and Su}]{miao2020diverse}
Shen-Yun Miao, Chao-Chun Liang, and Keh-Yih Su. 2020.
\newblock A diverse corpus for evaluating and developing english math word problem solvers.
\newblock In \emph{Proceedings of the 58th Annual Meeting of the Association for Computational Linguistics}, pages 975--984.

\bibitem[{Olsson(2009)}]{olsson2009literature}
Fredrik Olsson. 2009.
\newblock A literature survey of active machine learning in the context of natural language processing.

\bibitem[{OpenAI(2023)}]{openai2023gpt4}
OpenAI. 2023.
\newblock \href {http://arxiv.org/abs/2303.08774} {Gpt-4 technical report}.

\bibitem[{Pan et~al.(2023)Pan, Xing, Diao, Liu, Shum, Zhang, and Zhang}]{pan2023plum}
Rui Pan, Shuo Xing, Shizhe Diao, Xiang Liu, Kashun Shum, Jipeng Zhang, and Tong Zhang. 2023.
\newblock Plum: Prompt learning using metaheuristic.
\newblock \emph{arXiv preprint arXiv:2311.08364}.

\bibitem[{Pan et~al.(2024)Pan, Tian, Ding, Huang, Wen, and Li}]{pan2024pomp}
Shilong Pan, Zhiliang Tian, Liang Ding, Zhen Huang, Zhihua Wen, and Dongsheng Li. 2024.
\newblock \href {http://arxiv.org/abs/2401.05596} {Pomp: Probability-driven meta-graph prompter for llms in low-resource unsupervised neural machine translation}.

\bibitem[{Patel et~al.(2021)Patel, Bhattamishra, and Goyal}]{patel2021nlp}
Arkil Patel, Satwik Bhattamishra, and Navin Goyal. 2021.
\newblock Are nlp models really able to solve simple math word problems?
\newblock In \emph{Proceedings of the 2021 Conference of the North American Chapter of the Association for Computational Linguistics: Human Language Technologies}, pages 2080--2094.

\bibitem[{Pi et~al.(2022)Pi, Liu, Chen, Ziyadi, Lin, Gao, Fu, Lou, and Chen}]{pi2022reasoning}
Xinyu Pi, Qian Liu, Bei Chen, Morteza Ziyadi, Zeqi Lin, Yan Gao, Qiang Fu, Jian-Guang Lou, and Weizhu Chen. 2022.
\newblock Reasoning like program executors.
\newblock \emph{arXiv preprint arXiv:2201.11473}.

\bibitem[{Rae et~al.(2021)Rae, Borgeaud, Cai, Millican, Hoffmann, Song, Aslanides, Henderson, Ring, Young et~al.}]{rae2021scaling}
Jack~W Rae, Sebastian Borgeaud, Trevor Cai, Katie Millican, Jordan Hoffmann, Francis Song, John Aslanides, Sarah Henderson, Roman Ring, Susannah Young, et~al. 2021.
\newblock Scaling language models: Methods, analysis \& insights from training gopher.
\newblock \emph{arXiv preprint arXiv:2112.11446}.

\bibitem[{Raffel et~al.(2020)Raffel, Shazeer, Roberts, Lee, Narang, Matena, Zhou, Li, Liu et~al.}]{raffel2020exploring}
Colin Raffel, Noam Shazeer, Adam Roberts, Katherine Lee, Sharan Narang, Michael Matena, Yanqi Zhou, Wei Li, Peter~J Liu, et~al. 2020.
\newblock Exploring the limits of transfer learning with a unified text-to-text transformer.
\newblock \emph{J. Mach. Learn. Res.}, 21(140):1--67.

\bibitem[{Rotman and Reichart(2022)}]{rotman2022multi}
Guy Rotman and Roi Reichart. 2022.
\newblock Multi-task active learning for pre-trained transformer-based models.
\newblock \emph{Transactions of the Association for Computational Linguistics}, 10:1209--1228.

\bibitem[{Roy and McCallum(2001)}]{roy2001toward}
Nicholas Roy and Andrew McCallum. 2001.
\newblock Toward optimal active learning through sampling estimation of error reduction. int. conf. on machine learning.

\bibitem[{Scao et~al.(2022)Scao, Fan, Akiki, Pavlick, Ili{\'c}, Hesslow, Castagn{\'e}, Luccioni, Yvon, Gall{\'e} et~al.}]{scao2022bloom}
Teven~Le Scao, Angela Fan, Christopher Akiki, Ellie Pavlick, Suzana Ili{\'c}, Daniel Hesslow, Roman Castagn{\'e}, Alexandra~Sasha Luccioni, Fran{\c{c}}ois Yvon, Matthias Gall{\'e}, et~al. 2022.
\newblock Bloom: A 176b-parameter open-access multilingual language model.
\newblock \emph{arXiv preprint arXiv:2211.05100}.

\bibitem[{Schr{\"o}der et~al.(2022)Schr{\"o}der, Niekler, and Potthast}]{schroder2022revisiting}
Christopher Schr{\"o}der, Andreas Niekler, and Martin Potthast. 2022.
\newblock Revisiting uncertainty-based query strategies for active learning with transformers.
\newblock In \emph{Findings of the Association for Computational Linguistics: ACL 2022}, pages 2194--2203.

\bibitem[{Settles(2009)}]{settles2009active}
Burr Settles. 2009.
\newblock Active learning literature survey.

\bibitem[{Shum et~al.(2023)Shum, Diao, and Zhang}]{shum2023automatic}
Kashun Shum, Shizhe Diao, and Tong Zhang. 2023.
\newblock Automatic prompt augmentation and selection with chain-of-thought from labeled data.
\newblock In \emph{Findings of the Association for Computational Linguistics: EMNLP 2023}, pages 12113--12139.

\bibitem[{Si et~al.(2022)Si, Gan, Yang, Wang, Wang, Boyd-Graber, and Wang}]{si2022prompting}
Chenglei Si, Zhe Gan, Zhengyuan Yang, Shuohang Wang, Jianfeng Wang, Jordan Boyd-Graber, and Lijuan Wang. 2022.
\newblock Prompting gpt-3 to be reliable.
\newblock \emph{arXiv preprint arXiv:2210.09150}.

\bibitem[{Smith et~al.(2022)Smith, Patwary, Norick, LeGresley, Rajbhandari, Casper, Liu, Prabhumoye, Zerveas, Korthikanti et~al.}]{smith2022using}
Shaden Smith, Mostofa Patwary, Brandon Norick, Patrick LeGresley, Samyam Rajbhandari, Jared Casper, Zhun Liu, Shrimai Prabhumoye, George Zerveas, Vijay Korthikanti, et~al. 2022.
\newblock Using deepspeed and megatron to train megatron-turing nlg 530b, a large-scale generative language model.
\newblock \emph{arXiv preprint arXiv:2201.11990}.

\bibitem[{Talmor et~al.(2019)Talmor, Herzig, Lourie, and Berant}]{talmor2019commonsenseqa}
Alon Talmor, Jonathan Herzig, Nicholas Lourie, and Jonathan Berant. 2019.
\newblock Commonsenseqa: A question answering challenge targeting commonsense knowledge.
\newblock In \emph{Proceedings of the 2019 Conference of the North American Chapter of the Association for Computational Linguistics: Human Language Technologies, Volume 1 (Long and Short Papers)}, pages 4149--4158.

\bibitem[{Tay et~al.(2022)Tay, Dehghani, Tran, Garcia, Bahri, Schuster, Zheng, Houlsby, and Metzler}]{tay2022unifying}
Yi~Tay, Mostafa Dehghani, Vinh~Q Tran, Xavier Garcia, Dara Bahri, Tal Schuster, Huaixiu~Steven Zheng, Neil Houlsby, and Donald Metzler. 2022.
\newblock Unifying language learning paradigms.
\newblock \emph{arXiv preprint arXiv:2205.05131}.

\bibitem[{Touvron et~al.(2023{\natexlab{a}})Touvron, Lavril, Izacard, Martinet, Lachaux, Lacroix, Rozi{\`e}re, Goyal, Hambro, Azhar et~al.}]{touvron2023llama}
Hugo Touvron, Thibaut Lavril, Gautier Izacard, Xavier Martinet, Marie-Anne Lachaux, Timoth{\'e}e Lacroix, Baptiste Rozi{\`e}re, Naman Goyal, Eric Hambro, Faisal Azhar, et~al. 2023{\natexlab{a}}.
\newblock Llama: Open and efficient foundation language models.
\newblock \emph{arXiv preprint arXiv:2302.13971}.

\bibitem[{Touvron et~al.(2023{\natexlab{b}})Touvron, Martin, Stone, Albert, Almahairi, Babaei, Bashlykov, Batra, Bhargava, Bhosale et~al.}]{touvron2023llama2}
Hugo Touvron, Louis Martin, Kevin Stone, Peter Albert, Amjad Almahairi, Yasmine Babaei, Nikolay Bashlykov, Soumya Batra, Prajjwal Bhargava, Shruti Bhosale, et~al. 2023{\natexlab{b}}.
\newblock Llama 2: Open foundation and fine-tuned chat models.
\newblock \emph{arXiv preprint arXiv:2307.09288}.

\bibitem[{Wang et~al.(2022)Wang, Wei, Schuurmans, Le, Chi, and Zhou}]{wang2022self}
Xuezhi Wang, Jason Wei, Dale Schuurmans, Quoc Le, Ed~Chi, and Denny Zhou. 2022.
\newblock Self-consistency improves chain of thought reasoning in language models.
\newblock \emph{arXiv preprint arXiv:2203.11171}.

\bibitem[{Wei et~al.(2022{\natexlab{a}})Wei, Tay, Bommasani, Raffel, Zoph, Borgeaud, Yogatama, Bosma, Zhou, Metzler, Chi, Hashimoto, Vinyals, Liang, Dean, and Fedus}]{wei2022emergent}
Jason Wei, Yi~Tay, Rishi Bommasani, Colin Raffel, Barret Zoph, Sebastian Borgeaud, Dani Yogatama, Maarten Bosma, Denny Zhou, Donald Metzler, Ed~H. Chi, Tatsunori Hashimoto, Oriol Vinyals, Percy Liang, Jeff Dean, and William Fedus. 2022{\natexlab{a}}.
\newblock \href {https://openreview.net/forum?id=yzkSU5zdwD} {Emergent abilities of large language models}.
\newblock \emph{Transactions on Machine Learning Research}.
\newblock Survey Certification.

\bibitem[{Wei et~al.(2022{\natexlab{b}})Wei, Wang, Schuurmans, Bosma, Chi, Le, and Zhou}]{wei2022chain}
Jason Wei, Xuezhi Wang, Dale Schuurmans, Maarten Bosma, Ed~Chi, Quoc Le, and Denny Zhou. 2022{\natexlab{b}}.
\newblock Chain of thought prompting elicits reasoning in large language models.
\newblock \emph{arXiv preprint arXiv:2201.11903}.

\bibitem[{Xu et~al.(2024)Xu, Diao, Yang, and Wang}]{xu2024can}
Xin Xu, Shizhe Diao, Can Yang, and Yang Wang. 2024.
\newblock Can we verify step by step for incorrect answer detection?
\newblock \emph{arXiv preprint arXiv:2402.10528}.

\bibitem[{Xu et~al.(2021)Xu, Zhu, Wang, Sun, Cheng, Liu, Gao, He, Zeng, and Huang}]{xu2021human}
Yichong Xu, Chenguang Zhu, Shuohang Wang, Siqi Sun, Hao Cheng, Xiaodong Liu, Jianfeng Gao, Pengcheng He, Michael Zeng, and Xuedong Huang. 2021.
\newblock Human parity on commonsenseqa: Augmenting self-attention with external attention.
\newblock \emph{arXiv preprint arXiv:2112.03254}.

\bibitem[{Zelikman et~al.(2022)Zelikman, Mu, Goodman, and Wu}]{zelikman2022star}
Eric Zelikman, Jesse Mu, Noah~D Goodman, and Yuhuai~Tony Wu. 2022.
\newblock Star: Self-taught reasoner bootstrapping reasoning with reasoning.

\bibitem[{Zeng et~al.(2022)Zeng, Liu, Du, Wang, Lai, Ding, Yang, Xu, Zheng, Xia et~al.}]{zeng2022glm}
Aohan Zeng, Xiao Liu, Zhengxiao Du, Zihan Wang, Hanyu Lai, Ming Ding, Zhuoyi Yang, Yifan Xu, Wendi Zheng, Xiao Xia, et~al. 2022.
\newblock Glm-130b: An open bilingual pre-trained model.
\newblock \emph{arXiv preprint arXiv:2210.02414}.

\bibitem[{Zhang et~al.(2022{\natexlab{a}})Zhang, Roller, Goyal, Artetxe, Chen, Chen, Dewan, Diab, Li, Lin et~al.}]{zhang2022opt}
Susan Zhang, Stephen Roller, Naman Goyal, Mikel Artetxe, Moya Chen, Shuohui Chen, Christopher Dewan, Mona Diab, Xian Li, Xi~Victoria Lin, et~al. 2022{\natexlab{a}}.
\newblock Opt: Open pre-trained transformer language models.
\newblock \emph{arXiv preprint arXiv:2205.01068}.

\bibitem[{Zhang et~al.(2022{\natexlab{b}})Zhang, Zhang, Li, and Smola}]{zhang2022automatic}
Zhuosheng Zhang, Aston Zhang, Mu~Li, and Alex Smola. 2022{\natexlab{b}}.
\newblock Automatic chain of thought prompting in large language models.
\newblock \emph{arXiv preprint arXiv:2210.03493}.

\bibitem[{Zhou et~al.(2022)Zhou, Sch{\"a}rli, Hou, Wei, Scales, Wang, Schuurmans, Bousquet, Le, and Chi}]{zhou2022least}
Denny Zhou, Nathanael Sch{\"a}rli, Le~Hou, Jason Wei, Nathan Scales, Xuezhi Wang, Dale Schuurmans, Olivier Bousquet, Quoc Le, and Ed~Chi. 2022.
\newblock Least-to-most prompting enables complex reasoning in large language models.
\newblock \emph{arXiv preprint arXiv:2205.10625}.

\end{thebibliography}

\clearpage
\newpage
\appendix

\section{Experimental Settings}
\label{appendix:experimental_settings}
In this section, we describe the details of the datasets and evaluation metrics, the baseline models, and the implementation in the following three subsections.

\begin{table*}[h]
\centering
\footnotesize
\begin{tabular}{l|c|c|c|c|c|c}
\toprule
\textsc{Dataset} & \textsc{Task Type} & \textsc{\# Ex.} & \textsc{\# Train} & \textsc{\# Test} & \textsc{Eval. Split} & \textsc{Trans.} \\
\midrule
GSM8K~\citep{cobbe2021training} & Arithmetic & 8 & 7473 & 1319 & Test & \ding{55} \\
ASDiv~\citep{miao2020diverse} & Arithmetic  & 8 & - & 2096 & Test & \checkmark \\
SVAMP~\citep{patel2021nlp} & Arithmetic & 8 & - & 1000 & Test & \checkmark \\
AQuA~\citep{ling-etal-2017-program} & Arithmetic & 4 & 97467 & 254 & Test & \ding{55} \\
SingleEq~\citep{koncel-kedziorski-etal-2016-mawps} & Arithmetic & 8 & - & 508 & Test & \checkmark \\
CSQA*~\citep{talmor2019commonsenseqa}  & Commonsense & 7 & 9741 & 1221 & Dev & \ding{55} \\
StrategyQA*~\citep{10.1162/tacl_a_00370} & Commonsense & 6 & 2061 & 1880 & Dev & \ding{55} \\
Letter (4)~\citep{wei2022chain} & Symbolic & 4 & 1000 & 1000 & Test (OOD) & \ding{55} \\
\bottomrule
\end{tabular}
\caption{The statistics of the datasets used in this paper. 
\textsc{\# Ex.} are the number of few-shot chain-of-thought exemplars used to prompt each task in evaluation. 
\textsc{\# Train} and \textsc{\# Test} denote the number of training data and test data, respectively.
Note that in our experiments, we randomly sample 1000 data from the training set to reduce the computational cost and use the same test set as~\citet{wei2022chain}.
\textsc{Trans.}: A checkmark denotes that the exemplars are from other datasets and then transferred to this task. 
*: CSQA and StrategyQA do not have publicly available test set labels, so we simply follow the setting by~\citet{wei2022chain} to evaluate the performance of the development set.
}
\label{tab:dataset_statistic}
\vskip -1 em
\end{table*}

\subsection{Datasets and Evaluation Metrics}
\label{appendix:datasets-and-evaluation-metrics}

Following the standard evaluation settings in LLMs reasoning studies~\citep{wei2022chain}, our experiments are conducted on three types of datasets:
\begin{itemize}[leftmargin=*,label=$\bullet$,noitemsep,partopsep=0pt,topsep=0pt,parsep=0pt]
\item Arithmetic Reasoning: GSM8K~\citep{cobbe2021training}, ASDiv~\citep{miao2020diverse}, and SVAMP~\citep{patel2021nlp}, AQuA~\citep{ling-etal-2017-program}, and SingleEq~\citep{koncel-kedziorski-etal-2016-mawps}.
\item Commonsense Reasoning: CSQA~\citep{talmor2019commonsenseqa} and StrategyQA~\citep{10.1162/tacl_a_00370}.
\item Symbolic Reasoning: last letter concatenation~\citep{wei2022chain}.
This task evaluates the model's ability to concatenate the last letters of the words in a name.
The standard in-distribution setting is trivial, and previous methods have achieved almost 100\% accuracy~\citep{wei2022chain}.
We test on an out-of-distribution setting, where the prompts are two letters while the test questions are four letters.

\end{itemize}

The statistics of these datasets are reported in Table~\ref{tab:dataset_statistic}.
Note that in our experiments,
we randomly sample 1000 data from the training set to reduce the computational cost.
This may affect the performance of the uncertainty estimation.
Intuitively, more training data will help capture the data distribution, leading to more precise uncertainty estimation.
Given more financial support, the performance of our model will continue to increase.
To make a fair comparison, we use the same test set as~\citet{wei2022chain}.
We report the exact match accuracy as the evaluation metric.

\subsection{Baselines}
\label{appendix:baselines}
In our experiments, the following four methods serve as the main baselines:
\begin{itemize}[leftmargin=*,label=$\bullet$,noitemsep,partopsep=0pt,topsep=0pt,parsep=0pt]
    \item Chain-of-thought (CoT)~\citep{wei2022chain}: standard chain-of-thought prompting which provides four to eight human-written exemplars consisting of a series of intermediate reasoning steps.
    \item Self-consistency (SC)~\citep{wang2022self}: an improved version of CoT. Instead of greedy decoding, it samples a set of reasoning paths and chooses the most common answer. 
    \item Auto-CoT~\citep{zhang2022automatic}: 
    an automatic exemplar construction method by clustering and generating rationales with zero-shot prompting~\citep{kojima2022large}.
    \item Random-CoT: a baseline of {\ModelName}. 
    It shares the same annotation process as {\ModelName}. 
    The only difference is that it randomly samples questions from the training data for annotation instead of applying our proposed uncertainty metrics. 
\end{itemize}
Our experiments are mainly based on CodeX \texttt{code-davinci-002}~\citep{chen2021evaluating} for two reasons. 
First, it is the most capable model available at the time we were conducting our experiments, consistent with the observations in previous studies~\citep{wei2022chain, wang2022self, miao2020diverse}.
Second, it is free of charge in the initial limited beta period. 
In addition to \texttt{code-davinci-002}, we also test the performance with \texttt{text-davinci-002} and \texttt{text-davinci-003} to verify our method's effectiveness in the main experiment.
We call the APIs directly from OpenAI's services\footnote{\url{https://openai.com/api/}}.

\subsection{Implementation}
\label{sec:implementation}

\paragraph{Hyperparameters}
In our implementation, the model could only access the training data $D = \{X_{tr}, Y_{tr}\}$ before inference and is evaluated on the test data $D = \{X_{te}, Y_{te}\}$.
We apply the same number of exemplars as~\citet{wei2022chain}, which is 8 for GSM8K, ASDiv, SVAMP, and SingleEq, 7 for CSQA, 6 for StrategyQA, 4 for AQuA and Letter (4).
Given that some datasets (i.e., ASDiv, SVAMP, and SingleEq) only have the test split, we adopt the annotation result of GSM8K and transfer it to these datasets for inference.
The transfer details are in Table~\ref{tab:dataset_statistic}.
In the inference stage, we set temperature $T = 0.7$ and infer 40 times for each question. We then take the most consistent answer.

\paragraph{Uncertainty Estimation}
At this stage, we start with a few manually annotated exemplars to help infer answers in the uncertainty estimation stage.
These annotated exemplars are directly taken from~\citet{wei2022chain}.
We call it few-shot prompting trick to stabilize the prediction.
However, our method is not dependent on few-shot prompting, other exemplar-free methods like zero-shot prompting~\citep{kojima2022large} could be applied, and we demonstrate that it works well in Section~\ref{sec:ablation-few-shot}.
For the uncertainty metrics, we mainly report the performance of the disagreement-based ({\ModelName} (D)) and entropy-based ({\ModelName} (E)) methods.
Due to it having been observed that StrategyQA often ties with the maximum disagreement to be 2/2 = 1, we also take the frequency into consideration for {\ModelName} (D).

\paragraph{Annotation}
Our approach needs human annotation for a few selected questions.
The annotator is one of the co-authors and is familiar with machine learning and chain of thought prompting.
Owing to the focus of our method being the example selection, rather than the annotation, the annotator did not do trial and error and conduct the minimum human engineering, referring to the previous annotation practices~\citep{wei2022chain}.
Given a question, the annotator would mainly write the reasoning steps and give the true answer to it.
The effect of different annotators and the separate effects of selection and annotation are discussed in Sections~\ref{ablation}.

\section{Uncertainty Analysis}
\begin{figure*}[h]
    \centering
    \includegraphics[scale=0.75, trim=0 400 400 0,clip]{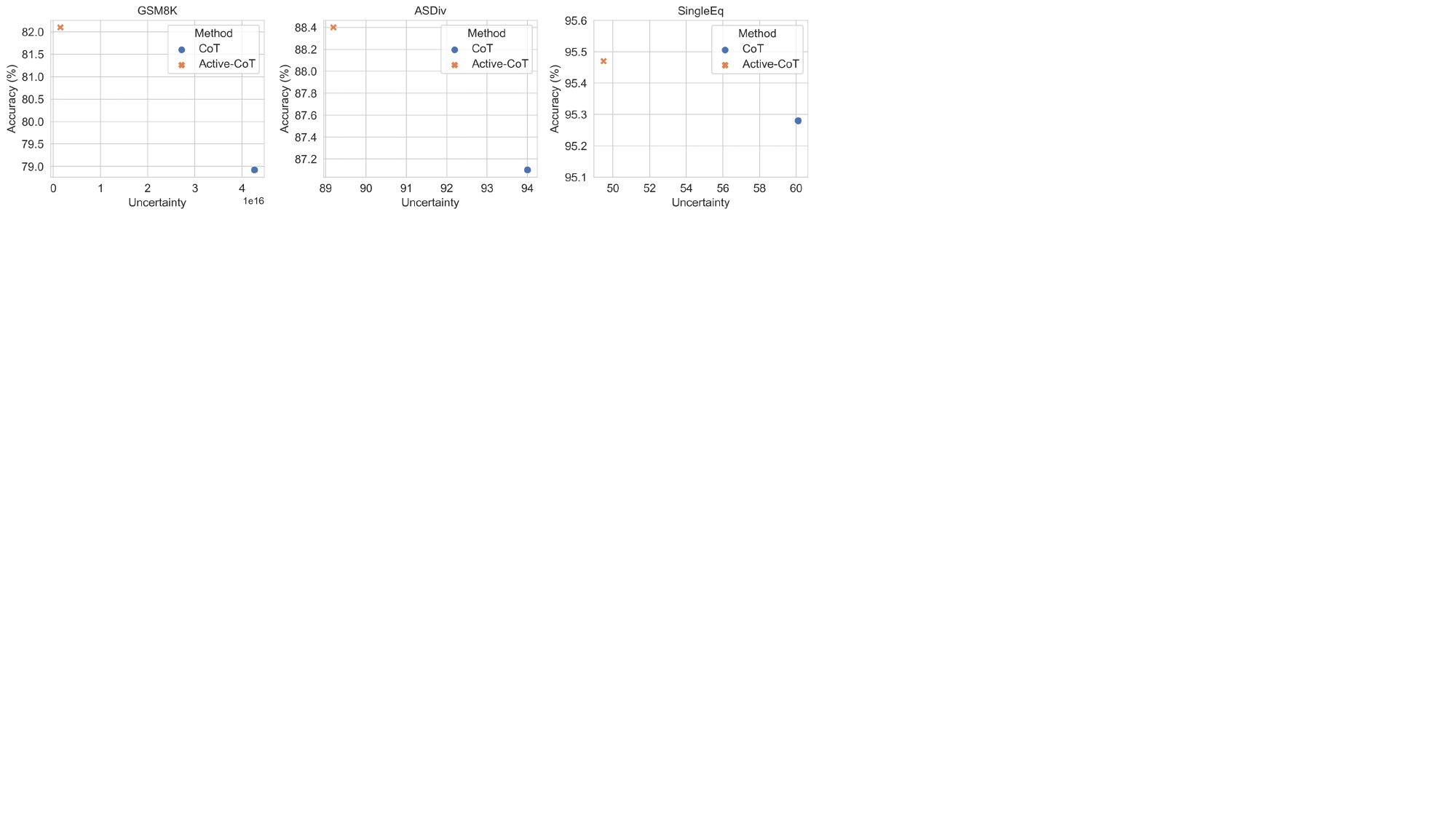}
    \caption{The relation between uncertainty and accuracy.}
    \label{fig:uncertainty_accuracy}
\end{figure*}

Figure~\ref{fig:uncertainty_accuracy} shows the relation between accuracy and uncertainty.

\section{Variance Analysis}
In our primary experiment, the step of uncertainty estimation necessitates querying each prompt in the training set $k$ times to assess uncertainty. 
However, for datasets with a large number of instances --- such as the GSM8K training set, which comprises 7,473 instances—to conserve resources --- we randomly sample 1,000 instances to estimate uncertainty. 
To expose the inherent randomness in this sampling process, we repeated the random sampling three times, aiming to examine its variance. 
The results, as illustrated in Table~\ref{tab:variance_analysis}, reveal that our method demonstrates robustness against the randomness of sampling. 
Sampling 1,000 instances proved to be sufficient for achieving stable and satisfactory results.

\begin{table}[h]
\centering
\footnotesize
\setlength{\tabcolsep}{4mm}{
\begin{tabular}{l|c|c|c}
\toprule
\textsc{Dataset} & Seed 1 & Seed 12 & Seed 42 \\
\midrule
GSM8K & 78.5 & 78.5 & 78.4 \\
\bottomrule
\end{tabular}
}
\caption{Experimental results on the GSM8K dataset with three seeds.
}
\label{tab:variance_analysis}
\end{table}


\section{Self-confidence-based Uncertainty Estimation}
\label{appendix:self-confidence}
Estimating the uncertainty can also be achieved by the LLMs themselves, namely self-confidence.
It can be obtained by querying the model with a manually crafted template $T$ like \textit{For the question $q$ and the predicted answer $a$,
report the confidence about the answer from choices. (a) very confident (b) confident (c) not confident (d) wrong answer}.
Then we select the least confident questions by:
\begin{equation}
    \begin{aligned}
    u &= \argmax_i(1 - \max_j P_\theta(a_j|q_i)) \\
      &= \argmin_i \max_j P_\theta(a_j|q_i),
    \end{aligned}
    \label{eq:self-confidence}
\end{equation}
where $P_\theta(a_j|q_i)$ is a categorical distribution from a set \{very confident, confident, not confident,  wrong answer\}.

\begin{table*}[h]
\centering
\footnotesize
\begin{tabular}{l|l}
\toprule
Question                                                   & \begin{tabular}[c]{@{}l@{}}Every 2 miles a car drives the tires rotate 725 times. Jeremy drives 400 miles a month. \\ If each tire can undergo 10,440,000 rotations how many years before the tire needs to be replaced?\end{tabular}                                                                                                                                                                                            \\ \midrule
\begin{tabular}[c]{@{}l@{}}Predicted\\ Answer\end{tabular} & \begin{tabular}[c]{@{}l@{}}Each tire can undergo 10,440,000 rotations. In 400 miles, the car travels 400 / 2 = 200 miles. \\ In 200 miles, each tire rotates 200 x 725 = 145,000 times. 145,000 / 10,440,000 = 0.0139. \\ The answer is 0.0139 years.\end{tabular}                                                                                                                                                                  \\ \midrule
Confidence                                                 & \begin{tabular}[c]{@{}l@{}}Report confidence about the answer from choices. \\
(a) very confident (b) confident (c) not confident (d) wrong answer \\ The answer is calculated by the number of rotations for each tire for a month and divide by \\ the maximum number of rotations it can do, which is 145,000/10,440,000=0.0139.\\ The logic is correct and no calculation error. So, choose (a). I am very confident.\end{tabular} \\ \midrule
Label     & 6   \\ \bottomrule
\end{tabular}
\caption{An example of self-confidence-based prompting process and the results.}
\label{tab:example-confidence}
\end{table*}

\section{Logits-based Uncertainty Estimation}

\begin{table}[h]
\centering
\footnotesize
\setlength{\tabcolsep}{1mm}{
\begin{tabular}{l|c|c|c|c}
\toprule
\textsc{Method} & GSM8K & AsDiv & SingleEq & Average \\
\midrule
CoT & 80.1 & 86.7 & 91.3 & 86.0 \\
{\ModelName} (D)  & 83.5 & 87.4 & 93.3 & 88.1 \\
{\ModelName} (E)  & 83.8 & 88.8 & 93.7 & 88.8 \\
{\ModelName} (Logits)  & 82.3 & 87.8 & 94.5 & 88.2 \\
\bottomrule
\end{tabular}
}
\caption{Comparison with logits-based uncertainty estimation methods. 
}
\label{tab:logit-based}
\end{table}

For models that provide logits, we can use the model’s output logits for uncertainty estimation. 
Therefore, we conduct further experiments to verify whether {\ModelName} still works.
We first conduct experiments with the logits returned by the gpt-3.5-turbo-0301 API. 
The results are shown in Table~\ref{tab:logit-based}. 
As we can see, using logits, the {\ModelName} method outperforms the traditional Chain of Thought (CoT), and is slightly better than the Disagreement-based method.

Secondly, we also conducted experiments using the logits from Llama-2-70b, but we found that Llama tends to exhibit overconfidence, leading to poorer results when using its logits as a measure of uncertainty. 
The phenomenon of overconfidence in the logits of deep neural networks has been discussed in previous works~\citep{guo2017calibration, kong2020calibrated, chen2022close}, and our observations are consistent with theirs. 
In the future, we plan to explore more methods of calibration so that logits can be used as a measure of uncertainty for active learning.

\section{Comparison with Diversity-based Methods}

\begin{table}[h]
\centering
\footnotesize
\setlength{\tabcolsep}{3mm}{
\begin{tabular}{l|c|c|c}
\toprule
\textsc{Method} & GSM8K & MultiArith & AddSub \\
\midrule
Auto-CoT & 62.8 & 93.2 & 91.9 \\
{\ModelName}  & \textbf{67.0} & \textbf{95.5} & \textbf{93.2} \\
\bottomrule
\end{tabular}
}
\caption{Comparison with Auto-CoT. 
The results of Auto-CoT are taken directly from the original paper. 
For a fair comparison, none of the results apply the self-consistency method.
{\ModelName} applies the rationales annotated by humans.
\textbf{Bold} represents the best among each dataset.
All the results are obtained with \texttt{code-davinci-002}.
}
\label{tab:comparison_autocot}
\end{table}

\begin{table}[t]
\centering
\footnotesize
\setlength{\tabcolsep}{1.2mm}{
\begin{tabular}{l|c|c|c|c}
\toprule
\textsc{Method} & GSM8K & AsDiv & SVAMP & SingleEq \\
\midrule
Complex-CoT & 76.3 & 82.4 & 79.9 & 93.3 \\
{\ModelName} (D)  & \textbf{77.1} & \textbf{83.6} & \textbf{85.5} & \textbf{96.0} \\
\bottomrule
\end{tabular}
}
\caption{Comparison with Complex-CoT. 
\textbf{Bold} represents the best among each dataset.
All the results are obtained with \texttt{gpt-3.5-turbo}.
}
\label{tab:comparison_complex_cot}
\end{table}

Auto-CoT~\citep{zhang2022automatic} proposes a diversity-based method for question selection, and ours proposes an uncertainty-based method for it.
In this section, we compare our method with Auto-CoT to demonstrate their effectiveness and differences.
Owing to Auto-CoT only reporting the results on GSM8K, MultiArith, and AddSub on \texttt{code-davinci-002} without self-consistency, we first compare our method with it on these three datasets in the same setting.
The results are shown in Table~\ref{tab:comparison_autocot}.
It is observed that {\ModelName} outperforms Auto-CoT by a large margin.
We attribute the improvement to uncertainty-based selection and human annotation.
Note that both diversity and uncertainty are useful for selecting the most informative questions, and they are complementary.
We consider the combination of diversity and uncertainty as an important future direction.

\section{Comparison with Complexity-based Methods}

\highlight{Complex-CoT~\citep{fu2022complexity} is a strong baseline which takes the complexity of prompts into consideration and proposes to select those complex prompts as exemplars.
We find that {\ModelName} outperforms Complex-CoT, demonstrating the effectiveness of our proposed uncertainty-based methods.
In addition, we can combine uncertainty and complexity to achieve better performance, and we leave this for future work.}

\section{Costs of {\ModelName}}

Compared with selecting questions by humans, our proposed method is more efficient. 
For a new task, users need to do trials and errors a lot of times which costs a lot of human effort with unstable performance. 
Even so, the selected questions are still suboptimal.
Second, as mentioned in Appendix~\ref{sec:implementation}, we limit the size of candidate instances to 1,000 which greatly reduces the cost. 
1,000 is a good balance between cost and performance. 
We verified that with more than 1,000 instances, the performance would converge. 
Doing uncertainty estimation 10 times with a pool of 1,000 questions is acceptable.
The cost is smaller than self-consistency, which usually requires 40 times inference, although it is an orthogonal technique and can be complementary to ours. 
In addition, inspired by the new experimental results in Section~\ref{appendix:transfer_gpt_llama}, we are excited to find that questions selected by smaller models (e.g., Llama) perform well with larger models (e.g., gpt-3.5-turbo). 
Considering models like Llama are open-source which does not cause API cost, one may use it (with GPU) to replace black-box API.

For the annotation, using human annotation is costly. 
We believe that using some techniques like zero-shot-CoT~\citep{kojima2022large} to replace manual annotation is a promising direction, and we will focus on exploring low-cost annotation methods in the future and integrate them with {\ModelName}.

\begin{table}[t]
\centering
\footnotesize
\setlength{\tabcolsep}{1.6mm}{
\begin{tabular}{l|c|c|c|c}
\toprule
\textsc{Method} & GSM8K & AsDiv & SVAMP & SingleEq \\
\midrule
Original CoT  & 74.2 & 82.5 & 83.8 & 95.0 \\
Longer CoT & 69.4 & 69.2 & 70.4 & 83.2 \\
{\ModelName} & 77.1 & 83.6 & 85.5 & 96.0 \\
\bottomrule
\end{tabular}
}
\caption{Ablation Study of Longer CoT Annotations. All the results are obtained with \tt{gpt-3.5-turbo}. 
}
\label{tab:longer_cot}
\end{table}

\section{Ablation Study of Longer CoT Annotations}
Furthermore, we conduct an ablation study to differentiate the impacts of longer CoT annotations from our method. 
To explore this, we extended the length of the original CoT~\citep{wei2022chain} annotations to an average of 155 words, comparable to our average length of 160 words. 
The results are shown in Table~\ref{tab:longer_cot}.
Our findings show that merely increasing the length of CoT annotations does not lead to improved performance, and in some cases, even reduces it. 
In contrast, our {\ModelName} method consistently demonstrates superior performance. 
This suggests that the selection of questions, rather than their length, contributes significantly to the improved results. Our approach effectively identifies and utilizes more informative examples for annotations.

\section{Full Exemplars Generated by {\ModelName}}
We display the full exemplars in Tables~\ref{tab:full_prompt_gsm8k_1}, \ref{tab:full_prompt_gsm8k_2}, \ref{tab:full_prompt_aqua}, \ref{tab:full_prompt_csqa}, \ref{tab:full_prompt_strategyQA}, \ref{tab:full_prompt_letter4}.

\begin{table*}[t]
\centering
\footnotesize
\begin{tabular}{p{15cm}}
\toprule
 \multicolumn{1}{c}{Exemplars}\\
\midrule
Q: As a freelancer, Baylor is paid for every finished work of a client he does on a freelance marketplace. Currently, he has \$4000 on his dashboard from previous work done. He is currently working for three clients, with the first client paying him half the amount of money he currently has on his dashboard once the job is done. The second client will pay him 2/5 times more money than the first client once Baylor finishes his work. The third client will pay him twice the amount of money the first and second clients pay him together once he finishes the job. How much money will Baylor have in his dashboard after all the clients pay him for his work? \\
A: \hl{Let's think step by step. To calculate how much money will Baylor have in his leaderboard after all clients pay him for this work, we need to know the amount of payment from each client and add them together with original value on the leaderboard. The first client will pay Baylor 1/2*\$4000=\$2000. The second client will pay Baylor 2/5 times more money than the first client, a total of 2/5*\$2000 = \$800. The total amount of money the second client will pay Baylor once the job is done is \$2000+\$800 = \$2800. The total amount of money the first and the second client will pay Baylor after their work is finished is \$2800+\$2000 = \$4800. The third client will pay him twice the amount of money the first and second clients pay him together, a total of 2*\$4800 = \$9600. Together, the three clients will pay Baylor \$9600+\$4800 = \$14400 once he finishes their jobs. In his dashboard, Baylor will have \$14400+\$4000 = \$18400.} The answer is 18400. \\
\\

Q: Josh and Anna were both born on August 17th, but in different years. To consolidate celebrations they also got married on August 17 when Josh turned 22. If today they're celebrating 30 years of marriage and their combined age is exactly 5 times what Josh's age was when they married, how old was Anna when they got married? \\
A: \hl{Let's think step by step. To calculate how old was Anna when they got married, we have to know their combined age, Josh's age after 30 years, and Anna's age after 30 years from their marriage. Since their combined age is 5 times Josh's age when he got married, their combined age is 5 * 22 = 110 years. Josh must be 30 years older than his age when they got married, so he is 22 + 30 = 52 years old now. Therefore, Anna's current age will be 110 - 52 = 58 years. If they married 30 years ago, Anna must have been 58 - 30 = 28 years old when they married} The answer is 28. \\
\\
Q: Mary is trying to get Zoom set up for the first time. She spent 10 minutes downloading the Mac version, only to realize she needed the Windows version, which took three times as long to download. During her first call, the audio glitched twice for 4 minutes each time and the video glitched once for 6 minutes. She spent twice as long talking without glitches as with glitches. How much time did Mary spend downloading Zoom and talking in total? \\
A: \hl{Let's think step by step. To calculate how much time did Mary spend downloading Zoom and talking in total, we need to know the downloading time of the Mac version, downloading time of the Windows version, the total time of audio glitches and video glitches, and the total glitch-free time. The Windows version's downloading time is three times that of the Mac version, which is 3 * 10 = 30 minutes. The total time of audio glitches and video glitches is 4 minutes/glitch * 2 audio glitches + 6 minutes audio glitches = 14 minutes. The glitch-free time is twice as long as the time with glitch, which is 2 * 14 = 28 minutes. The total time Mary spent downloading Zoom and talking is to add all together, 10 minutes + 30 minutes + 14 minutes + 28 minutes =  82 minutes.} The answer is 82. \\
\\

Q: Valerie needs to put stamps on the envelopes she is about to mail. She has thank you cards for each of her grandmother, uncle and aunt for the birthday presents they sent. She also has to pay the water bill and the electric bill separately. She wants to send three more mail-in rebates than she does bills and she has twice as many job applications as rebates to mail. How many stamps does she need if everything needs 1 stamp except the electric bill, which needs 2? \\
A: \hl{Let's think step by step. To calculate how many stamps does Valerie needs to put on the envelopes she is about to mail, we need to know the number of thank you cards for birthday presents, the number of bills, the number of mail-in rebates, the number of job applications, and the total number of mails to send. Since there are three people to thank, the number of thank you cards is 3. There are two bills which are the water bill and the electric bill. The number of mail is 3 more than the number of bills, which is 3 + 2 = 5. The number of job applications is twice as many as main-in rebates, which is 2 * 5 = 10 applications. The total number of mails to send is 3 + 2 + 5 + 10 = 20 pieces of mail to send. Each mail needs 1 stamp but electric bills, which need 2 stamps. So she needs 20 + 1 = 21 stamps.} The answer is 21. \\
\\

Q: At the duck park, there are 25 mallard ducks and ten less than twice as many geese as ducks. Then, a small flock of 4 ducks arrived at the park and joined the birds already there. If five less than 15 geese leave the park, how many more geese than ducks remain at the park? \\
A: \hl{Let's think step by step. To calculate how many more geese than ducks remain at the park, we need to know the number of geese initially at the park, the number of ducks after 4 ducks arrived, the number of geese left, and the number of geese remains. The number of geese initially at the park is ten less than twice as many geese as ducks, which is 2*25 - 10 = 40 geese. The number of ducks in the park after 4 ducks arrived is 25 + 4 = 29 ducks. The number of geese left is 5 less than 15 geese which are 15 - 5 = 10 geese left. The number of geese remaining is 40 - 10 = 30 geese remaining. Therefore, there are 30 - 29 = 1 more geese than ducks remaining at the park.} The answer is 1. \\
\\

\bottomrule
\end{tabular}
\caption{Exemplars for full chain of thought prompt selected and annotated from GSM8K. 
This set of exemplars is used by GSM8K, ASDiv, SVAMP, and SingleEq.}
\label{tab:full_prompt_gsm8k_1}
\end{table*}

\begin{table*}[t]
\centering
\footnotesize
\begin{tabular}{p{15cm}}
\toprule
\multicolumn{1}{c}{Exemplars}\\
\midrule

Q: John decides to stop delivering the newspapers he is supposed to deliver and instead steals them to recycle them for cash. The Monday-Saturday papers weigh 8 ounces each. The Sunday paper weighs twice as much. He is supposed to deliver 250 papers a day. He doesn't deliver them for ten weeks. If one ton of paper recycles for \$20, how much did he make? \\
A: \hl{Let's think step by step. To calculate how much did John make if one ton of paper was recycled for \$20, we need to know the weight of a weekly delivery of the papers in pounds, the total weight of paper delivered in 10 weeks, and the total weight of paper in tons. The weekly delivery of a paper consists of 6 days of 8 ounces of paper and one day of 2*8 = 16 ounces of paper, so the weekly weight of paper is 6 * 8  + 16 = 64 ounces, which is 64/16=4 pounds. Since John needs to deliver 250 papers a day, the total weight of paper delivered in a week is 4 * 250 = 1000 pounds. The total weight of papers in 10 weeks is 10 * 1000 = 10000 pounds. The weight of paper in tons is 10000/2000 = 5 tons. So he makes 5 * 20 = \$100 from recycling the paper.} The answer is 100. \\
\\

Q: John buys a chair. He then buys a table that is 3 times the price of the chair. Then, he buys a couch that is 5 times the price of the table. If John paid \$380 for all these items, what is the price of the couch? \\
A: \hl{Let's think step by step. To calculate the price of the couch, we need to know the price of the chair, the price of the table, and the relation between the chair, table, couch, and total money paid. Let x be the price of the chair, 3*x be the price of the table, and 5*(3*x) = 15*x be the price of the couch. The relationship between the chair, table, couch, and the total price paid is x + 3*x + 15*x = \$380, which is 19*x = 380, and x=20. The price of the couch is 15*x, which is 15*20 = \$300.} The answer is 300. \\
\\

Q: A hay farmer harvested 560 bales of hay from 5 acres of grass per month last year. This year, he planted an additional 7 acres of grass. If the farmer also owns 9 horses and each horse consumes 3 bales of hay a day, how many bales of hay would the farmer have left by the end of December if he starts feeding them this year's hay beginning the first day of September? \\
A: \hl{Let's think step by step. To calculate how many bales of hay would the farmer have left by the end of December if he starts feeding the horse this year's hay beginning the first day of September, we need to know the number of bales of hay that can harvest from each acre of grass, the number of acres of grass the farmer has this year, the total number of bales of hay can harvest per month, the number of acres of grass the farmer has this year, the total number of bales of hay can harvest this year, the number of days to feed the horse from September to December, the number of bales of hay his house eats per day, and the total number of bales of hay his houses will eat. The number of bales of hay that can harvest from each acre of grass is 560/5 = 112 bales of hay each month. The number of acres of grass the farmer has this year is 7 + 5 = 12 acres of grass. The number of bales of hay that can harvest per month from the 12 acres of grass is 12 * 112 = 1344 bales of hay per month. The total number of bales of hay he can harvest this year is 1344 * 12 = 16128. The number of days to feed the horse from September to December is a total of 30 + 31 + 30 + 31 = 122 days. The number of bales of hay his horse eats each day is 3 * 9 = 27. The total number of bales of hay his horses consumed in 122 days is 27 * 122 = 3294. The total number of bales remaining is 16128-3294 = 12834.} The answer is 12834. \\
\\

\bottomrule
\end{tabular}
\caption{(Cont.) Exemplars for full chain of thought prompt selected and annotated from GSM8K. 
This set of exemplars is used by GSM8K, ASDiv, SVAMP, and SingleEq.}
\label{tab:full_prompt_gsm8k_2}
\end{table*}

\begin{table*}[t]
\centering
\footnotesize
\begin{tabular}{p{15cm}}
\toprule
\multicolumn{1}{c}{Exemplars}\\
\midrule
Q: As a bicycle salesperson, Norman earns a fixed salary of \$20 per week plus \$6 per bicycle for the first six bicycles he sells, \$12 per bicycle for the next six bicycles he sells, and \$18 per bicycle for every bicycle sold after the first 12. This week, Norman earned more than twice as much as he did last week. If he sold A bicycles last week and Y bicycles this week, which of the following statements must be true? I. y > 2x II. y > x III. y > 3 Answer Choices: (A) I only (B) II only (C) I and II (D) II and III (E) I, II, and III \\
A: \hl{Let's think step by step. This is kind of like doing your taxes, lets summarize the data in terms of bike sales and salary. 0 Bikes - \$20. 6 Bikes - \$56. 12 Bikes - \$128. More than 12 bikes - 128 + 18*(a-12) where x is the total number of bikes sold (when a>12). A = cycles sold last week. AS = Salary last week. Y = cycles sole this week. YS = Salary this week. given YS>2AS. Let's test all statements with A=0;. AS = 20. YS >40. True when Y = 4. satisfies all the statements but we can only be sure of iii as Y will only increase from here. So iii must be true. Eliminate A,B,C. lets test all statements with A=1;. AS = 26. YS>52. True when Y=6. Still satisfies all the statements - Nothing achieved. lets test all statements with A=6. AS = 56. YS>112. True when Y = 11. This proves statement i wrong, hence eliminate E so we are left with D.} The answer is D. \\
\\

Q: If k is a non-negative integer and 15\^k is a divisor of 823,435 then 5\^k - k\^5 = Answer Choices: (A) 0 (B) 1 (C) 35 (D) 120 (E) 245 \\
A: \hl{Let's think step by step. 8+2+3+4+3+5 = 25, so this number is not divisible by 3 and thus not divisible by 15. Therefore, k=0. 5\^k - k\^5 =1-0=1.} The answer is B. \\
\\

Q: For the past 3 weeks, Rikki has gone to the gym an average of 2 times per week. After 1 more week, the average number of times Rikki goes to the gym increases to 3 times per week. How many times did Rikki visit the gym during the most recent week? Answer Choices: (A) 7 (B) 3 (C) 6 (D) 2 (E) 4 \\
A: \hl{Let's think step by step. Total times per week over 3 weeks = 3*2 = 6. Total times per week over 4 weeks = 4*3 = 12. Number of visits in most recent week = 12-6 = 6 visits.} The answer is C. \\
\\

Q: A clock is set right at 8 a.m. The clock gains 10 minutes in 24 hours. What will be the true time when the clock indicates 1 p.m. on the following day? Answer Choices: (A) 48 minutes past 12 (B) 48 minutes past 32 (C) 48 minutes past 19 (D) 48 minutes past 11 (E) 48 minutes past 16 \\
A: \hl{Let's think step by step. Time from 8 a.m. on a day to 1 p.m. on the following day = 29 hours. 24 hours 10 minutes of this clock = 24 hours of the correct clock. Hours of this clock = 24 hours of the correct clock. 29 hours of this clock = hours of the correct clock = 28 hrs 48 minutes of the correct clock. Therefore, the correct time is 28 hrs 48 minutes after 8 a.m. This is 48 minutes past 12.} The answer is A. \\
\\

\bottomrule
\end{tabular}
\caption{Exemplars for full chain of thought prompt selected and annotated from AQuA.}
\label{tab:full_prompt_aqua}
\end{table*}

\begin{table*}[t]
\centering
\footnotesize
\begin{tabular}{p{15cm}}
\toprule
\multicolumn{1}{c}{Exemplars}\\
\midrule
Q: Where would you find a sloth that is not afraid of being hunted? Answer Choices: (A) forest canopy (B) nature preserve (C) zoo (D) tropical rainforest (E) commercial \\
A: \hl{Let's think step by step. To answer the question of where you would find a sloth that is not afraid of being hunted, we need to know where sloths live and when sloths are not afraid of being hunted. Sloths live in places where their habitats are well protected. Sloths are not afraid of being hunted when hunting is illegal in their living places. Of the above choices, (B) nature preserve makes the most sense because it well protects sloths' habitats and forbids hunting.} The answer is B. \\
\\

Q: What gift do people claim that god has for people? Answer Choices: (A) bible (B) everywhere (C) judge people (D) care for people (E) work miracles \\
A: \hl{Let's think step by step. To answer the question of what gift do people claim that god has for people, we need to know what is a gift from god and what people cannot do without the gift from god. The gift from god is an ability that people normally cannot achieve. People normally cannot create miracles unless it is gifted by god. Of the above choices, (E) work miracles make the most sense because miracles refer to something that people cannot achieve unless it is gifted by god.} The answer is E. \\
\\

Q: What is probably not built out of a wooden rod? Answer Choices: (A) broom (B) broom closet (C) old clock (D) hardware store (E) old houses \\
A: \hl{Let's think step by step. To answer the question of what is probably not built out of a wooden rod, we need to know what is a wooden rod and what a wooden rod can do. A wooden rod is a stick made of wood of various sizes. A wooden rod can be used as a part of tools or furniture. Of the above choices, (E) old houses make the most sense because the wooden rod is too weak to support old houses.} The answer is E. \\
\\

Q: Where do you use condoms? Answer Choices: (A) supermarket (B) dick (C) bedroom (D) medicine chest (E) bed \\
A: \hl{Let's think step by step. To answer the question of where do you use condoms, we need to know when do people need condoms and where is the place people stay when they use condoms. People use condoms when they have sex. People have sex and use condoms in their bedrooms. Of the above choices, (C) bedroom makes the most sense because it is most polite and appropriate for people to have sex with condoms in the bedroom.} The answer is C. \\
\\

Q: What happens soon after Christmas? Answer Choices: (A) buy presents for others (B) summer (C) easter (D) see work (E) halloween \\
A: \hl{Let's think step by step. To answer the question of what happens soon after Christmas, we need to know what means soon after Christmas and what can be happened soon after Christmas. Soon after Christmas means some days after Christmas day. Normally another holiday happens soon after Christmas. Of the above choices, (C) easter makes the most sense because easter is a holiday with the closest date after Christmas day.} The answer is C. \\
\\

Q: Finding information requires what other than research materials? Answer Choices: (A) internet (B) ask questions (C) time (D) searching for (E) logic \\
A: \hl{Let's think step by step. To answer the question of finding information requires what other than research materials, we need to know what people do when they are finding information and what are research materials. People spend time looking up something with purpose is called finding information. Research materials normally refer to the stuff they are looking for and necessary tools such as the internet. Of the above choices, (C) time makes the most sense because people definitely need to spend time to find information.} The answer is C. \\
\\

Q: Before someone can adopt the parent must do what with their offspring? Answer Choices: (A) biological child (B) give away (C) birth (D) abandon (E) orphan \\
A: \hl{Let's think step by step. To answer the question of before someone can adopt the parent must do what with their offspring, we need to know what is the requirement to adopt and what is the legal action to do with their offspring. Normally people can adopt when they don't have offspring or give up having their offspring. It is legal to decide not to have babies, which is giving away their offspring. Of the above choices, (B) give away makes the most sense because people must give away their offspring to be able to adopt.} The answer is B. \\
\\

\bottomrule
\end{tabular}
\caption{Exemplars for full chain of thought prompt selected and annotated from CommonsenseQA.}
\label{tab:full_prompt_csqa}
\end{table*}

\begin{table*}[t]
\centering
\footnotesize
\begin{tabular}{p{15cm}}
\toprule
\multicolumn{1}{c}{Exemplars}\\
\midrule
Q: Is it possible to get killed walking to the Very Large Telescope? \\
A: \hl{Let's think step by step. The Very Large Telescope is in the Atacama Desert. The Atacama Desert is the driest hot desert in the world. Thus, it is possible to get killed walking in the Atacama desert.} The answer is yes. \\
\\

Q: Did Clark Gable marry more women once than Richard Burton? \\
A: \hl{Let's think step by step. Richard Burton was married to four different women. Clark Gable was married to five different women, one time each. Thus, Clark Gable married more women than Richard Burton.} The answer is yes. \\
\\

Q: Can a human heart last from NYC to Raleigh NC by Toyota Hiux? \\
A: \hl{Let's think step by step. Human hearts can last up to 6 hours outside the body. The distance from NYC to Raleigh, NC is 505 miles. The top speed of a Toyota Hilux is 105 miles per hour. The time needed to travel from NYC to Raleigh, NC is 505 miles / 105 miles per hour = 4.8 hours. Thus, the time needed to travel from NYC to Raleigh NC is less than 6 hours.} The answer is yes. \\
\\

Q: Is H's most common two letter pair partner a freebie in Wheel of Fortune bonus round? \\
A: \hl{Let's think step by step. H forms the most common two letter pair in the English language along with the letter T. The Wheel of Fortune bonus round gives the player six free letters: R, S, T, L, N, E. Thus, The letter T is inclued in the given six free letters in the Wheel of Fortune bonus round.} The answer is yes. \\
\\

Q: Could someone in Tokyo take a taxi to the The Metropolitan Museum of Art? \\
A: \hl{Let's think step by step. Tokyo is located in Japan. The Metropolitan Museum of Art is in the United States. Japan and the United States are separated by the Pacific Ocean. A taxi is not capable of travelling over the Pacific Ocean. Thus, someone in Tokyo cannot take a taxi to the The Metropolitan Museum of Art.} The answer is no. \\
\\

Q: Does Olympia Washington share name with Hephaestus's workshop location? \\
A: \hl{Let's think step by step. Olympia Washington is named after Mount Olympus, which is a mountain range in Washington named after the ancient Greek Mount Olympus. Hephaestus was the ancient Greek god of the forge and had a workshop on Mount Olympus. Thus, Olympia Washington share name with Hephaestus's workshop location} The answer is yes. \\
\\

\bottomrule
\end{tabular}
\caption{Exemplars for full chain of thought prompt selected and annotated from StrategyQA.}
\label{tab:full_prompt_strategyQA}
\end{table*}

\begin{table*}[t]
\centering
\footnotesize
\begin{tabular}{p{15cm}}
\toprule
\multicolumn{1}{c}{Exemplars}\\
\midrule
Q: Take the last letters of the words in ``Willie Valeria Zane'' and concatenate them. \\
A: \hl{Let's think step by step. The last letter of ``Willie'' is ``e''. The last letter of ``Valeria'' is ``a''. The last letter of ``Zane'' is ``e''. Concatenating them is ``eae''.} The answer is eae. \\
 \\

Q: Take the last letters of the words in ``Kylie Guille Lakshmi'' and concatenate them. \\
A: \hl{Let's think step by step. The last letter of ``Kylie'' is ``e''. The last letter of ``Guille'' is ``e''. The last letter of ``Lakshmi'' is ``i''. Concatenating them is ``eei''.} The answer is eei. \\
\\

Q: Take the last letters of the words in ``Carmelo Reinaldo Sade'' and concatenate them. \\
A: \hl{Let's think step by step. The last letter of ``Carmelo'' is ``o''. The last letter of ``Reinaldo'' is ``o''. The last letter of ``Sade'' is ``e''. Concatenating them is ``ooe''.} The answer is ooe. \\
\\

Q: Take the last letters of the words in ``Gabe Ventura Woody'' and concatenate them. \\
A: \hl{Let's think step by step. The last letter of ``Gabe'' is ``e''. The last letter of ``Ventura'' is ``a''. The last letter of ``Woody'' is ``y''. Concatenating them is ``eay''.} The answer is eay. \\
\\

\bottomrule
\end{tabular}
\caption{Exemplars for full chain of thought prompt selected and annotated from Letter (4).}
\label{tab:full_prompt_letter4}
\end{table*}

\end{document}